%% file: main.tex
\documentclass[10pt,twocolumn,letterpaper]{article}
\usepackage[pagenumbers]{cvpr}
\input{preamble}
\definecolor{linkblue}{rgb}{0.21,0.49,0.74}
\usepackage[breaklinks,colorlinks,allcolors=linkblue]{hyperref}
\hypersetup{pdftitle={Movie2Game-Bench: Holistic Evaluation of Film-to-Game Adaptation Beyond Playability},pdfauthor={Jialing Xie, Yifan Zhou, Yiwen Zhang, Junjun He}}
\title{Movie2Game-Bench: Holistic Evaluation of\\Film-to-Game Adaptation Beyond Playability}
\author{%
Jialing Xie$^{1,*}$ \quad
Yifan Zhou$^{2,4,*}$ \quad
Yiwen Zhang$^{1}$ \quad
Junjun He$^{3,4,\dagger}$\\[4pt]
{\normalsize $^{1}$Independent \quad $^{2}$Shanghai Jiao Tong University}\\
{\normalsize $^{3}$Shanghai Innovation Institute \quad $^{4}$Ant Group}\\[4pt]
{\small $^{*}$Equal contribution. \quad $^{\dagger}$Corresponding author.}%
}
\begin{document}
\maketitle
\input{sec/0_abstract}
\input{figures/fig1_overview}
\input{sec/1_intro}
\input{sec/2_related_work}
\input{sec/3_benchmark}

\input{sec/4_evaluation}

\input{sec/5_results}

\input{sec/6_conclusion}
\FloatBarrier
\phantomsection
\label{main:end}
\section*{LLM Usage Statement}
Large language models are the evaluated game generators and serve as source-reference extractors, evaluators, and GUI model players, as described in the protocol. OpenAI Codex assisted with paper planning, drafting, local data aggregation, and LaTeX preparation; GPT-6-Astra assisted with draft review. The reported experimental scores come from the existing evaluation records. Figures combine supplied illustrations of generation and evaluation with plots of recorded statistics.
{\small
\bibliographystyle{ieeenat_fullname}
\bibliography{main}
}
\clearpage
\appendix
\setcounter{table}{0}
\renewcommand{\thetable}{S\arabic{table}}
\setcounter{figure}{0}
\renewcommand{\thefigure}{S\arabic{figure}}
\onecolumn
\raggedbottom
\input{sec/appendix_contents}
\clearpage
\input{sec/A_experimental_setup}
\clearpage
\input{sec/B_evaluation_protocols}
\input{sec/B_scoring_prompts}
\clearpage
\input{sec/C_additional_results}
\clearpage
\input{sec/D_game_gallery}
\end{document}

%% file: preamble.tex
\usepackage[T1]{fontenc}
\usepackage{amsmath,amssymb,booktabs,multirow,array,tabularx}
\usepackage{graphicx,xspace,enumitem,placeins}
\usepackage{tikz}
\usepackage{tcolorbox}
\tcbuselibrary{breakable}
\usetikzlibrary{arrows.meta,positioning,calc}
\newcommand{\bench}{M2G-bench\xspace}
\newcommand{\fullbench}{Movie2Game-Bench\xspace}

\definecolor{generalblue}{HTML}{E9F0FC}
\definecolor{moviegreen}{HTML}{E9F3EC}
\definecolor{playorange}{HTML}{FFF1DE}
\definecolor{inkblue}{HTML}{263C55}
\definecolor{levelzero}{HTML}{D8DFE8}
\definecolor{levelone}{HTML}{7BA0CF}
\definecolor{leveltwo}{HTML}{376C76}
\definecolor{LighterGray}{HTML}{F7F7FA}
\definecolor{DeepPurple}{HTML}{4A326F}
\newtcolorbox{scoringprompt}[1]{
  breakable,
  title={#1},
  colback=LighterGray,
  colframe=DeepPurple,
  colbacktitle=DeepPurple,
  coltitle=white,
  fonttitle=\bfseries,
  fontupper=\small,
  boxrule=0.6pt,
  arc=2pt,
  left=9pt,right=9pt,top=7pt,bottom=7pt,
  before skip=12pt,after skip=12pt,
  title after break={#1 (continued)}
}
\newcommand{\promptheading}[1]{\par\medskip\noindent\textbf{\emph{\textcolor{DeepPurple}{#1}}}\par\smallskip}
\newcolumntype{Y}{>{\raggedright\arraybackslash}X}

%% file: sec/0_abstract.tex
\begin{abstract}
Adapting a film into a game requires selecting source elements and translating them into player actions, rules, and feedback. We introduce Movie2Game-Bench (M2G-bench), a benchmark for evaluating film-to-game adaptation beyond playability. M2G-bench combines general evaluation, movie-specific evaluation, and playtesting by three GUI model players. The protocol connects source content to gameplay steps and implementation evidence, capturing both content preservation and interactive realization. The benchmark comprises 22 films and 658 source-specific reference items. Evaluating 14 models on the same films with their selected primary harnesses reveals complementary strengths in game quality, adaptation, and player experience. A shared weakness concerns playable sequences: this category has the lowest content-presence mean in all 14 models, averaging 37.9 out of 100 compared with 54.8 for iconic elements. Among judgments of retained emotional-development content, narrative choices account for 60.0\% and scenario operation for 7.2\%. These findings identify two linked priorities for film-to-game adaptation: preserving source situations that can support gameplay and connecting retained content to player actions and consequences.
\end{abstract}

%% file: figures/fig1_overview.tex
\begin{figure}[!t]
\centering
\includegraphics[width=\columnwidth]{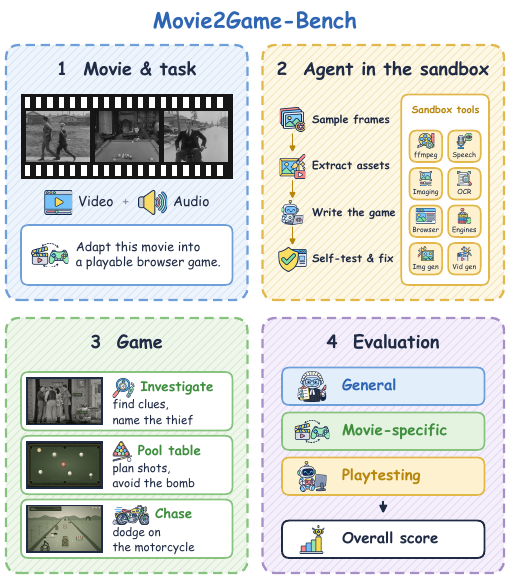}
\caption{\textbf{Movie2Game-Bench overview.} A coding agent designs, builds, and tests a browser game from a film. General Evaluation, Movie-specific Evaluation, and Playtesting assess the resulting adaptation.}
\label{fig:overview}
\end{figure}

%% file: sec/1_intro.tex
\section{Introduction}
\label{sec:intro}
A film can become a detective game, a driving challenge, or an interactive narrative. Each adaptation makes a choice about what the player should do with the source material. Recognizing a character, reproducing a location, and turning a conflict into a consequential action are distinct achievements. Evaluating film-to-game generation therefore requires connecting the source film to the generated work and to the player's experience of that work.

Recent coding agents can construct complete games, and game-generation benchmarks assess functionality, visual quality, and player experience~\cite{gamecraft,gamexpert,gamedev,playtest}. These advances establish a foundation for studying adaptation. A film, however, supplies narrative events and audiovisual situations rather than an executable specification of game rules. The adaptation problem is to select meaningful source content and realize it through a playable design. The same scene can support multiple valid mechanics, so evaluation needs a source reference that accommodates differences in implementation.

We introduce \fullbench (\bench) for holistic evaluation of film-to-game adaptation (\cref{fig:overview}). \emph{General Evaluation} assesses Playability, Gameplay, and Presentation. \emph{Movie-specific Evaluation} distinguishes source-content presence from interactive realization. \emph{Playtesting} records three GUI model players' actions and experience. Together, these perspectives reveal how a game functions, what it preserves from the film, and how players engage with it.

The protocol traces source content into player interaction. Film-specific reference items cover plot, character relationships, iconic elements, playable sequences, and emotional development. Evaluators connect each item's realization to game content, gameplay steps, and implementation evidence. Retaining a chase as text and letting a player control movement through it express different relationships to the source. The protocol records this difference while accommodating narrative choices, puzzles, and action mechanics.

\bench comprises 22 films and 658 source-specific reference items. We evaluate 14 models in their selected primary harnesses, averaging each score over all 22 films with zero for unsuccessful generation. Astra has the highest General Evaluation, Playtesting, and Overall means; Gemini has the highest Movie-specific mean. All 14 models preserve playable sequences less fully than the other four reference categories, identifying a shared gap in retaining source situations that could ground gameplay. Item-level analysis then examines how retained plot, playable sequences, and emotional development are expressed through different forms of player participation.

Our contributions are:
\begin{itemize}[leftmargin=*,topsep=3pt,itemsep=2pt]
\item A full-film adaptation benchmark with 22 films and 658 source-specific reference items, supporting open game designs under a common browser-game task.
\item A holistic evaluation protocol that connects general game quality, movie-specific adaptation, and playtesting through traceable source, code, and gameplay evidence.
\item Evidence of a shared content-preservation gap in playable sequences across all 14 models, complemented by analysis of the operation forms used to express retained film content.
\end{itemize}

%% file: sec/2_related_work.tex
\section{Related Work}
\label{sec:related}
\paragraph{Interactive worlds and game creation.}
Genie learns controllable environments from unlabelled video~\cite{genie}; GameNGen simulates a real-time game engine through diffusion~\cite{gamengen}; and GameGen-X generates game videos with interactive control~\cite{gamegenx}. Film adaptation adds a semantic design problem: reorganizing a source narrative around player actions in an executable game.

GameDevBench evaluates tutorial-derived development tasks~\cite{gamedev}, and OpenGameEval checks edited scenes and simulated play in Roblox Studio~\cite{opengameeval}. GameCraft-Bench evaluates complete game construction through runtime replay, mechanics, content, and presentation~\cite{gamecraft}. CEGame-Bench covers five engines with real-input play verification~\cite{cegame}. Harness-of-Harness organizes coding agents into iterative planning, coding, and testing loops, including game development~\cite{hoh}. These works establish executable artifacts and observed behavior as evaluation targets. \bench adds a full film as the source for designing content and interaction.

\paragraph{Game quality and playtesting.}
GameXpert-Bench covers game generation, repair, and optimization, including visual quality and player experience~\cite{gamexpert}. GameASG-Bench uses declared interfaces and executable checks to assess generated software behavior~\cite{gameasg}. \bench combines a general game-quality profile with evaluation of the adaptation's relationship to its source film.

Orak and LMGame-Bench evaluate game-playing agents across diverse games and examine agent modules~\cite{orak,lmgame}. PlaytestArena uses independent GUI judges to test expected in-play behavior, while Play2Code incorporates playtester feedback into iterative game creation~\cite{playtest}. Our GUI players produce trajectories, screenshots, reports, and questionnaires that support game-quality and experience scores. Movie-specific Evaluation connects those interactions to source events and relationships.

\paragraph{Source reconstruction, replication, and adaptation.}
Video2Game reconstructs interactive environments from video using explicit scene representations and simulation~\cite{video2game}. CineRecon evaluates editable dynamic-scene reconstruction from cinematic references~\cite{cinerecon}. Film adaptation permits changes in viewpoint, temporal organization, and interaction style, motivating a reference inventory of meaningful source content.

GameReplica studies the recovery of visual and behavioral properties of reference games~\cite{gamereplica}, and SWE-Game evaluates development tasks derived from existing games, including reference-video conditions~\cite{swegame}. Such games supply target mechanics to reproduce; films supply events from which mechanics must be devised. RPGAgent coordinates narrative, spatial, and mechanical generation to turn story synopses into playable role-playing prototypes~\cite{rpgagent}. \bench evaluates multiple systems conditioned on complete audiovisual films through a common source-content inventory and actual gameplay evidence.

\paragraph{Long-video understanding and content use.}
MovieChat addresses long-video understanding through memory~\cite{moviechat}; LongVideoBench studies long-context video-language understanding~\cite{longvideobench}; and Video-MME evaluates multimodal video understanding~\cite{videomme}. Film-to-game adaptation makes downstream content use observable in a persistent artifact: characters enter the game world and events become playable situations. M2G-bench evaluates this use through source-grounded content judgments, observed interaction, and player experience.

%% file: sec/3_benchmark.tex
\section{Movie2Game-Bench}
\label{sec:benchmark}
\subsection{Task and Source Reference}
The task is to adapt a complete audiovisual film into a self-contained browser game. The instruction leaves the game type and design to the generator. Each agent works unattended for up to three hours, with local development tools and model-service access. The game consists of an HTML entry point and bundled assets, served by a static local server during evaluation. The environment provides no application backend or general Internet access. All systems receive the same task instruction and artifact contract.

The film collection contains 22 titles spanning different periods, languages, and audiovisual forms, including silent films, dialogue-centered dramas, animation, and action films. The collection supports different adaptation opportunities: investigating evidence, coordinating characters, controlling movement, and organizing resources. These opportunities motivate the film selection while leaving the generated gameplay open. The complete film list and reference-item counts appear in \cref{tab:films}; screenshots of generated adaptations appear in \cref{app:gallery}.

Each film is associated with a fixed inventory of source-specific reference items. The 658 items cover five categories: plot structure, characters and relationships, iconic elements, playable sequences, and emotional development. Most films have 30 items; two have 29. An item names content to check in an adaptation and includes source grounding. It does not prescribe a unique interface or mechanic. For example, an evidence-related plot item can be expressed through inspecting objects, comparing testimony, or choosing a line of inquiry.

Reference construction uses timestamped frame grids, visual records, and dialogue transcripts. Six independent extractions, organized as generalist and specialist roles, identify candidate items. A merger groups candidates that address the same content within a category and selects representative statements. Timestamp checks and dialogue verification ground the resulting inventory in the film. The inventory is fixed before the adaptation judgments and enters the evaluation stage separately from the generator's instruction. This gives all systems a common source reference while preserving their freedom to choose a playable design.

\subsection{Primary Systems}
We evaluate 14 models, listed in \cref{tab:main}. Claude Opus and Sonnet use Claude Code~\cite{claude_code} as their primary harness; GPT-6-Astra and GPT-6.1-Sol use Codex~\cite{codex_cli}. The other models use an existing adaptation of DeepSeek Harness~\cite{deepseek_harness}. A harness includes the agent's tool interface, prompt organization, context management, and execution loop. The primary assignments follow the existing experimental setup.

Every primary system is evaluated on the same 22 films. For each model--harness--film setting, the recorded artifact is the latest successful artifact when one exists; an unsuccessful game generation receives zero in every reported score. Artifact selection uses execution status rather than evaluation scores. This produces a common film-level comparison of the selected systems.

\Cref{app:harness} additionally compares all 14 models across three harnesses on the same eight films, holding the model and film fixed and scoring unsuccessful generations as zero.

%% file: sec/4_evaluation.tex
\section{Evaluation Protocol}
\label{sec:evaluation}
\input{figures/fig_evaluation_dimensions}
\input{figures/fig2_pipeline}
The evaluation dimensions in \cref{fig:evaluation_dimensions} are connected through recorded evidence (\cref{fig:pipeline}). Three GUI model players explore the game, generating observations that evaluators can trace to actions and implementation. Movie-specific Evaluation additionally receives the film-derived reference inventory. Each perspective produces a score on a 0--100 scale.

\subsection{General Evaluation}
General Evaluation assesses the generated game through 15 dimensions organized into three groups. \emph{Playability} covers onboarding, goals, recovery, pacing, challenge, and interaction feedback. \emph{Gameplay} covers functioning mechanics, choices and consequences, gameplay variety, narrative integration, content completeness, and narrative presentation. \emph{Presentation} covers visual production, interface clarity, and visual design. The groups contain six, six, and three dimensions, respectively.

The rubric contains 103 questions with explicit decision rules. Three evaluators answer 102 questions from gameplay records, screenshots, and applicable implementation evidence. One question, which checks the first effective state change, is answered programmatically. Question-level majority decisions provide the inputs to a dimension-specific grading rule.

The grading rule checks any defined minimum requirements before considering progressively stronger criteria at levels 3 to 5. A failure of a minimum requirement assigns level 1. Otherwise, grading starts at level 2 and advances through the defined levels in order: each level requires a mean question pass value of at least 0.5 and passage of every applicable mandatory question. Grading stops at the first unmet level; an unverified minimum requirement caps the result at level 3. Experience-related dimensions first pool each question's answers across valid player sessions, and assign level 1 only when the same minimum requirement fails in every scored session. The detailed rules and a mechanics example appear in \cref{app:rubric}.

Let $\ell_d\in\{1,2,3,4,5\}$ be the assigned level of dimension $d$. We linearly map the endpoints 1 and 5 to 0 and 100, giving $100(\ell_d-1)/(5-1)$. Let $\mathcal D_k$ contain the dimensions with an assigned numeric level in group $k$; dimensions recorded as inapplicable, untriggered, or undetermined are excluded. Groups $k=1,2,3$ denote Playability, Gameplay, and Presentation. We average within each group, then across the three groups:
\begin{equation}
\begin{aligned}
G_k&=\frac{1}{|\mathcal D_k|}\sum_{d\in\mathcal D_k}100\frac{\ell_d-1}{5-1},\\
G&=\frac{1}{3}\sum_{k=1}^{3}G_k.
\end{aligned}
\label{eq:general}
\end{equation}
The level-to-score mapping yields $0,25,50,75,100$. Aggregation gives equal weight to the three groups despite their different numbers of dimensions. The main table reports all three group scores and their General Total.

\input{tables/main_results}
\subsection{Movie-specific Evaluation}
Movie-specific Evaluation asks what source content appears in the game and how the player participates in it. For every reference item, evaluators separately record content presence and interactive realization. Presence has three levels: absent, partly present, and accurately retained. Interactive realization also has three levels:
\begin{enumerate}[leftmargin=*,topsep=3pt,itemsep=2pt,label=\arabic*.,start=0]
\item Passive presentation or confirmation-based progression.
\item Choice-based participation, or a generic mechanic weakly related to the source action.
\item Direct operation of a source-related situation with corresponding feedback, including mechanics strongly aligned with the target action.
\end{enumerate}
These correspond to numeric labels $0,1,2$. Interaction is assessed for plot, playable sequences, and emotional development. Character relationships and iconic elements use the content-presence judgment. Separately, evaluators record the operation form: narrative choice, weakly or strongly aligned generic mechanism, or scenario operation. This descriptive field does not enter scoring; passive or confirmation-based progression denotes level 0. The highest interaction level includes both scenario operation and strongly aligned generic mechanics.

\paragraph{Grounding in observed play.}
A highest-level interaction judgment must point to an existing gameplay step and quote implementation evidence found in the game's code. The recorded step establishes that the interaction occurred during play, and the code helps identify its mechanism. The evidence bundle includes reports, screenshots, game code, and programmatic source-asset traces. Generation plans and intended features are excluded from the evaluator's evidence. An item that is not triggered, one that lacks sufficient evidence, and one confirmed absent receive distinct statuses.

For a concrete example, the \emph{F1} reference inventory includes a late-race sequence involving driving, a pit stop, repair, and recovery. The recorded Astra adaptation includes controllable driving and a sprint sequence. Both movie-specific evaluators identify these actions while recording the missing repair and rescue components. The item-level judgment captures partial preservation and actual interaction within the same source situation. This is the purpose of keeping content presence and interactive realization separate.

\paragraph{Scoring source realization.}
Normalize presence and interaction labels to $p,i\in\{0,\frac12,1\}$. For an item in an interaction-bearing category, its realization score is
\begin{equation}
q=p(0.1+0.9i).
\label{eq:interaction}
\end{equation}
For characters and relationships or iconic elements, $q$ is proportional to content presence. Let $\bar q_c$ be the mean score within category $c$, $w_c$ its weight, and $q_c^{\max}$ its maximum attainable score. Each evaluator divides the weighted score by its weighted maximum:
\begin{equation}
M=100\cdot\frac{\sum_c w_c\bar q_c}{\sum_c w_c q_c^{\max}}.
\label{eq:movie}
\end{equation}
This normalization maps a fully realized inventory to 100. The category rules, weights, and maxima are listed in \cref{tab:movie_scoring}. Two evaluators score independently, and their resulting Movie-specific scores are averaged. Content fidelity averages category-level presence with the same five weights. Interactive realization averages the three applicable category-level interaction means using their category weights, renormalized over their sum. Both diagnostics are scaled to 0--100 and averaged across evaluators.

An untriggered presence or interaction field receives zero and remains in its applicable mean's denominator. Its recorded status remains distinct from confirmed absence. Observed fields retain their assigned values: when presence is known but interaction is untriggered, the content contribution remains and interaction is zero. Unknown interaction is excluded from interaction and realization means; known presence still contributes to content fidelity. Each game's score depends on its own judgments and the fixed film inventory. Interaction distributions use raw numeric labels for present content.

\subsection{Playtesting}
Three GUI model-player roles independently explore each game through screenshots and mouse or keyboard actions. The configured players are DeepSeek-V4.1-Flash, Gemini-4-Argon, and GPT-6.1-Sol. They receive the screen and interaction interface rather than privileged access to internal game state. Each session has a budget of 400 steps and 180 minutes, with stopping and replay rules specified in \cref{app:playtesting}.

After playing, each model writes a report and completes a 24-item experience questionnaire. The items cover onboarding and control, gameplay systems, content and presentation, and adaptation-related interaction. The questionnaire evaluates what the player actually experienced, including visible consequences, pacing, and the relation between narrative and action. The same session also produces screenshots and action traces for the other evaluators.

For each player, the valid questionnaire ratings are averaged on a one-to-five scale. The player means are then aggregated and mapped to 0--100 using $25(\bar r-1)$. This produces the Playtesting score $P$. The four groups organize the questionnaire; aggregation uses valid items within each player. Ten human participants also completed 100 playthroughs using this questionnaire (\cref{app:human_playtesting}).

\input{figures/fig_content_preservation}
\subsection{Aggregation and Coverage}
A game's Overall score combines the three perspectives:
\begin{equation}
O=w_GG+w_MM+w_PP.
\label{eq:overall}
\end{equation}
Here $w_G$, $w_M$, and $w_P$ are the weights for General Evaluation, Movie-specific Evaluation, and Playtesting, respectively. We use $w_G=0.4$, $w_M=0.4$, and $w_P=0.2$ for every game and system.
Game-level scores are rounded to one decimal place and averaged over the same 22 films. Unsuccessful game generation receives zero in the three General groups, all three perspective totals, and Overall. The film-level denominator therefore includes every assigned adaptation task.

The main table reports means. Score intervals and model-pair differences (\cref{app:estimates}) use 10,000 joint bootstrap resamples of the 22 films. Every system shares the sampled film indices, so a pairwise comparison retains the same film composition. These resamples hold game generation and scoring fixed. Harness comparisons report means over the same eight films.

%% file: figures/fig_evaluation_dimensions.tex
\par\medskip\noindent
\begin{minipage}{\columnwidth}
\centering
\includegraphics[width=0.84\columnwidth]{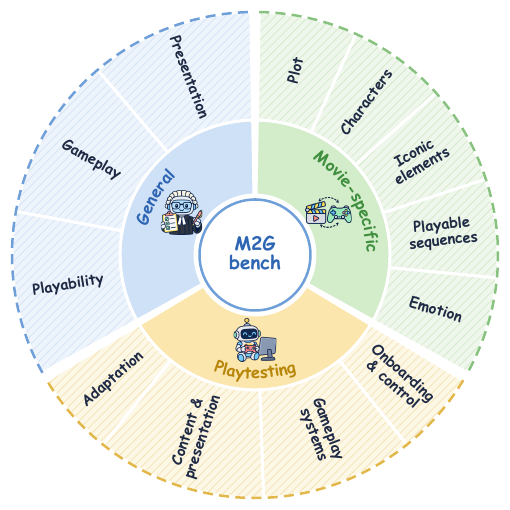}
\captionof{figure}{\textbf{Evaluation dimensions} across the three perspectives.}
\label{fig:evaluation_dimensions}
\end{minipage}
\par\medskip

%% file: figures/fig2_pipeline.tex
\begin{figure*}[!t]
\centering
\includegraphics[width=\textwidth]{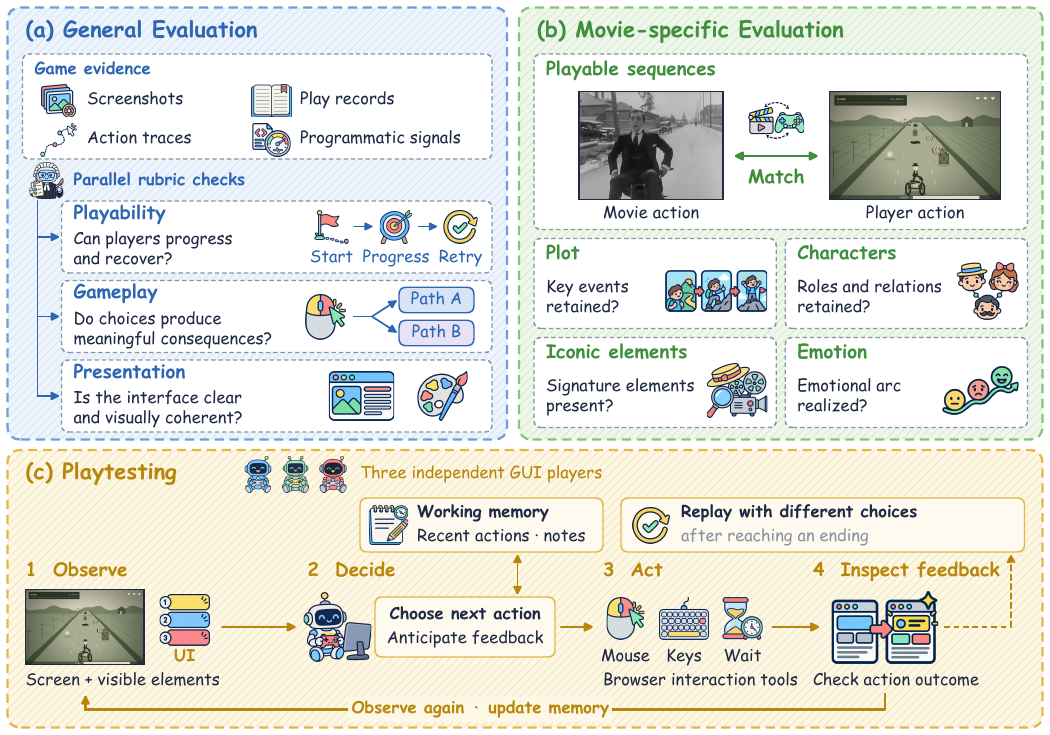}
\caption{\textbf{Evaluation framework.} General Evaluation measures Playability, Gameplay, and Presentation. Movie-specific Evaluation traces film content into game content and interaction. Three GUI model players provide gameplay evidence and experience questionnaires for Playtesting. The three perspectives form the benchmark score profile and Overall score.}
\label{fig:pipeline}
\end{figure*}

%% file: tables/main_results.tex
\begin{table*}[t]
\caption{\textbf{Results in each model's selected primary harness.} All scores are means over the same 22 films (0--100, higher is better); unsuccessful game generation receives zero in every column. Bold marks column maxima at the displayed precision, including ties. CC: Claude Code; DSH: DeepSeek Harness.}
\label{tab:main}
\centering\small
\setlength{\tabcolsep}{4.4pt}
\begin{tabular*}{\textwidth}{@{\extracolsep{\fill}}llrrrrrrr@{}}
\toprule
\multirow{2}{*}{Model} & \multirow{2}{*}{Harness} & \multicolumn{4}{c}{General Evaluation} & \multirow{2}{*}{\shortstack{Movie-specific\\Evaluation}} & \multirow{2}{*}{Playtesting} & \multirow{2}{*}{Overall} \\
\cmidrule(lr){3-6}
 & & Playability & Gameplay & Presentation & Total & & & \\
\midrule
GPT-6-Astra & Codex & \textbf{86.3} & \textbf{86.7} & \textbf{72.3} & \textbf{82.0} & 26.9 & \textbf{59.0} & \textbf{55.4} \\
GPT-6.1-Sol & Codex & 81.8 & 81.0 & 65.9 & 76.3 & 32.7 & 58.3 & 55.2 \\
Kimi-K3.1-Pro & DSH & 78.6 & 65.7 & 64.8 & 69.7 & 27.3 & 54.7 & 49.7 \\
HY4 & DSH & 68.5 & 51.5 & 61.4 & 60.4 & 33.0 & 52.1 & 47.8 \\
Claude Opus 5.5 & CC & 60.2 & 65.1 & 47.7 & 57.7 & 33.4 & 49.8 & 46.4 \\
DeepSeek-V4.1-Flash & DSH & 68.3 & 50.6 & 62.1 & 60.3 & 29.0 & 48.8 & 45.5 \\
Grok-4.7 & DSH & 59.2 & 52.3 & 59.9 & 57.1 & 30.6 & 49.6 & 45.0 \\
Gemini-4-Argon & DSH & 55.2 & 40.9 & 53.8 & 50.0 & \textbf{36.0} & 46.3 & 43.6 \\
Claude Sonnet 5.5 & CC & 60.3 & 54.2 & 45.8 & 53.6 & 30.1 & 45.4 & 42.6 \\
Muse-Wonder-1.0 & DSH & 63.1 & 44.9 & 42.4 & 50.1 & 27.4 & 45.5 & 40.1 \\
Qwen3.8-Max & DSH & 56.1 & 46.5 & 43.9 & 48.8 & 23.6 & 42.2 & 37.4 \\
Mimo-V2.6-Pro & DSH & 47.7 & 36.0 & 43.9 & 42.6 & 26.3 & 37.9 & 35.1 \\
MiniMax-M3.1-Flash & DSH & 48.1 & 32.0 & 47.7 & 42.7 & 18.9 & 38.9 & 32.4 \\
Step-5-Preview & DSH & 46.4 & 42.7 & 39.4 & 42.9 & 17.0 & 38.3 & 31.6 \\
\bottomrule
\end{tabular*}
\end{table*}

%% file: figures/fig_content_preservation.tex
\begin{figure}[t]
\centering
\includegraphics[width=\columnwidth]{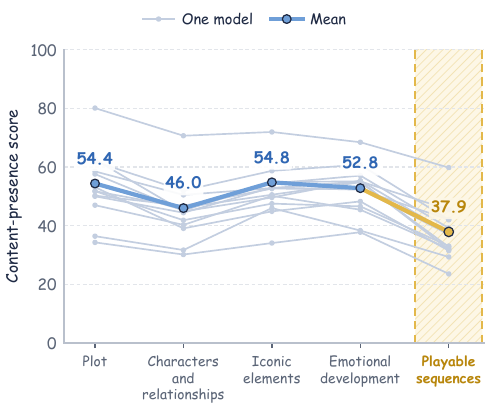}
\caption{\textbf{Preservation of source-film content.} Each thin line connects one model's category means over the same 22 films; the thick line averages models equally. Unsuccessful generation receives zero. Playable sequences, highlighted in yellow, have the lowest content-presence mean within every model.}
\label{fig:content_preservation}
\end{figure}

%% file: sec/5_results.tex
\section{Results and Analysis}
\label{sec:results}
\subsection{Complementary System Strengths}
\Cref{tab:main} compares 14 primary systems on the same 22 films. Astra has the highest General Evaluation mean at 82.0 and Playtesting mean at 59.0. Gemini has the highest Movie-specific mean at 36.0. Astra and Sol achieve Overall means of 55.4 and 55.2, respectively, through different combinations of game quality and source-specific realization.

Kimi ranks third in General and Playtesting and ninth in Movie-specific Evaluation; Gemini ranks first in Movie-specific Evaluation, tenth in General, and eighth in Playtesting. These profiles distinguish broad game quality from adaptation-specific achievement, exposing capabilities that an Overall ranking alone would obscure.

\paragraph{Evaluation consistency.}
The two Movie-specific evaluators correlate at $r=0.91$ (\cref{app:consistency}). Across 100 blind human playthroughs by ten participants, mean model-player scores correlate with human scores at Spearman $\rho=0.60$ and average 11.1 points higher. Participants rated the games without access to model-player scores (\cref{app:human_playtesting}).

\subsection{Preservation of Source Content}
\Cref{fig:content_preservation} compares content presence across the five source-reference categories. Each thin line follows one model's means over the same 22 films, with zero for unsuccessful generation. Playable sequences have the lowest mean within every one of the 14 models. Averaging models equally gives 37.9 for playable sequences, compared with 54.4 for plot, 54.8 for iconic elements, 52.8 for emotional development, and 46.0 for characters and relationships.

Playable-sequence references describe concrete source situations that can support player actions. Their lower presence scores identify a shared weakness in preserving these situations in the generated adaptation. Plot and recognizable film elements receive greater coverage, while the sequences that could ground gameplay are less fully retained. \Cref{fig:content_preservation} locates this gap at the level of content presence. \Cref{fig:interaction} then examines the operation forms used for content that is retained, connecting what enters an adaptation to how players engage with it. \Cref{tab:adaptation_profiles} provides the aggregate content and interaction diagnostics.

\subsection{How Source Content Reaches Players}
\input{figures/fig3_interaction}
\input{figures/fig_tool_design}
\input{figures/fig4_efficiency}
\Cref{fig:interaction} distinguishes the operation forms through which retained film content reaches players. Narrative choices account for 44.3\% of plot judgments, 40.1\% of playable-sequence judgments, and 60.0\% of emotional-development judgments. Scenario operation accounts for 12.6\%, 20.1\%, and 7.2\%, respectively. These are frequencies among retained-content judgments pooled across both evaluators; the operation forms describe implementation independently of the interaction score.

Among these three categories, emotional development relies most heavily on narrative choices, while playable sequences have the largest share of scenario operation. Connecting source situations to player actions and consequential feedback provides a concrete direction for richer film-to-game adaptation.

\subsection{Tool Use and Player-facing Design}
Realizing source situations also requires readable information and operable controls. \Cref{fig:tool_design} compares the ten primary systems that share DeepSeek Harness. DeepSeek allocates 66.8\% of native calls to Shell, compared with Kimi's 25.2\%; the direction holds on all 22 films. The common interface therefore supports distinct implementation strategies for the open adaptation task.

The game-design panel complements these process profiles with player-facing outcomes. Kimi leads this cohort in mechanic repertoire (70.5) and interface clarity (48.9), while DeepSeek leads art craftsmanship (78.4). All ten systems score higher in visual design than interface clarity under the respective rubrics. The latter checks readable information, accessible controls, and visible goals and state, identifying concrete priorities for making generated adaptations easier to navigate.

\subsection{Generation Cost and Efficiency}
\Cref{fig:cost_pareto} compares mean generation cost with Overall for the same primary systems and 22 films. Cost includes model and media generation; unsuccessful runs retain their incurred cost and receive zero Overall. MiniMax, DeepSeek, Sol, and Astra form the observed Pareto frontier, spanning complementary cost--quality operating points.

Sol achieves 55.2 Overall at \$1.129 per film; Astra achieves 55.4 at \$3.925. Sol's mean generation cost is 71.2\% lower, with an Overall mean 0.11 points below Astra's before rounding. DeepSeek achieves 45.5 for \$0.782. Astra leads in speed and recorded token efficiency, averaging 16.2 minutes and 1.38 million cumulative tokens per film. \Cref{tab:efficiency} provides the cost, token, and runtime breakdown, exposing efficiency advantages beyond quality alone.

%% file: figures/fig3_interaction.tex
\begin{figure}[!t]
\centering
\includegraphics[width=\columnwidth]{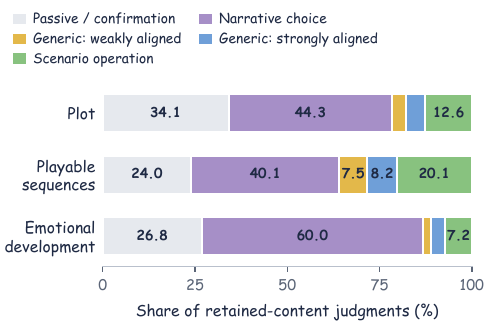}
\caption{\textbf{How film content becomes player interaction.} Operation forms for plot, playable sequences, and emotional development. We pool both evaluators' judgments with presence $>0$ and numeric interaction across primary-setting games, normalizing within each category (2,503, 3,326, and 1,877 judgments).}
\label{fig:interaction}
\end{figure}

%% file: figures/fig_tool_design.tex
\begin{figure*}[!t]
\centering
\includegraphics[width=\textwidth]{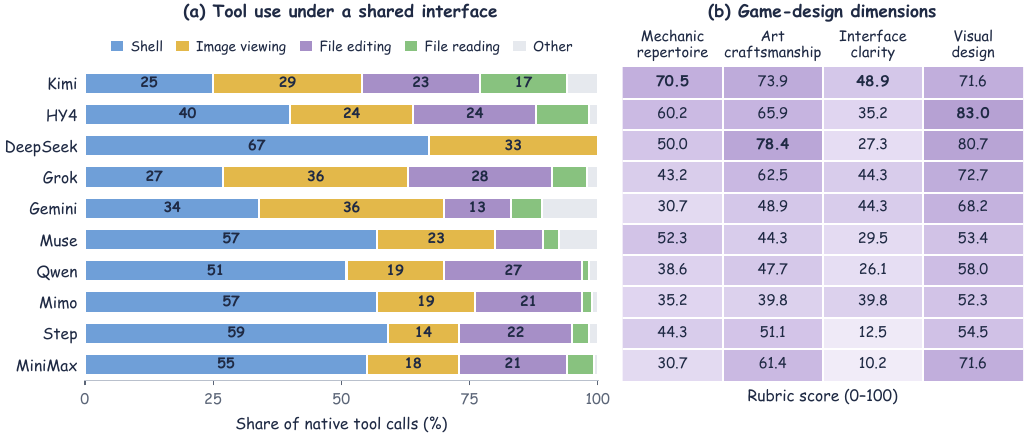}
\caption{\textbf{Tool use and player-facing design.} Ten primary DeepSeek Harness systems on the same 22 films. (a) Native tool-call shares, computed within each film and averaged equally across films. Shell also supports file operations and testing. (b) Game-design rubric means, with zero for unsuccessful generation. Bold marks each column's highest displayed mean within this cohort.}
\label{fig:tool_design}
\end{figure*}

%% file: figures/fig4_efficiency.tex
\begin{figure*}[t]
\centering
\includegraphics[width=\textwidth]{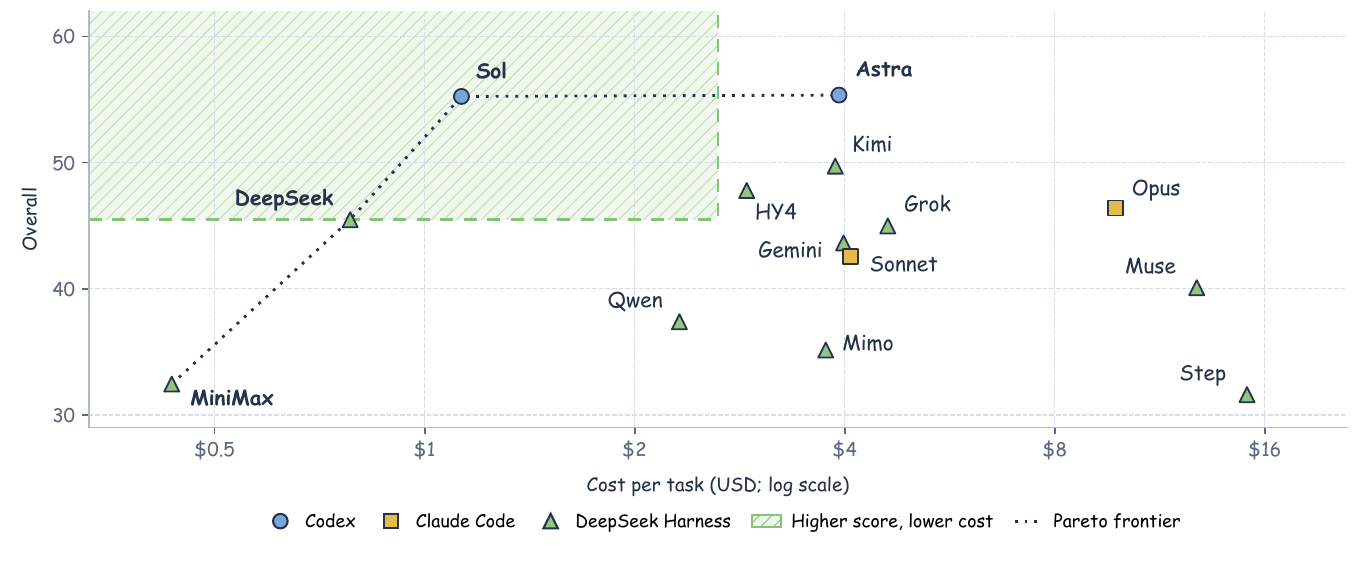}
\caption{\textbf{Overall versus cost per task.} Each point is a model in its selected primary harness, averaged over 22 films. Cost includes model and media generation. Colors and marker shapes identify harnesses; the green quadrant indicates higher score and lower cost. The dotted line connects the observed Pareto frontier: MiniMax, DeepSeek, Sol, and Astra. Detailed costs, tokens, and runtimes appear in \cref{tab:efficiency}.}
\label{fig:cost_pareto}
\end{figure*}

%% file: sec/6_conclusion.tex
\section{Conclusion}
\label{sec:conclusion}
\bench evaluates game quality, source-specific adaptation, and player experience. Across 22 films, these perspectives reveal complementary model strengths. Playable sequences have the lowest content-presence mean within all 14 models, identifying a shared gap in preserving source situations that can support gameplay. Retained emotional content is most often conveyed through narrative choices. Preserving source situations and connecting retained content to consequential player actions emerge as linked priorities for film-to-game adaptation.

Tool-use profiles reveal distinct implementation strategies under a shared interface. Readable information, accessible controls, and visible goals are design priorities, while the cost--Overall frontier provides complementary deployment choices. Together, these analyses connect source understanding, game construction, and player experience. The current inventory emphasizes source coverage and interaction; transformative adaptations can pursue additional artistic goals.

\paragraph{Future work.}
Future directions include generation methods that explicitly map source-film sequences to player actions, game states, and consequential feedback. Extending evaluation to longer games and additional engines offers a way to study richer interaction structures. Broader human playtesting can further assess how these mechanics convey source-film emotions and support creative reinterpretation, connecting content preservation to sustained player participation.

%% file: sec/appendix_contents.tex
\section*{Supplementary Material}
\phantomsection
\pdfbookmark[0]{Supplementary Contents}{supp:contents}
\begin{center}\large\bfseries Contents\end{center}
\medskip
\newcommand{\suppentry}[2]{%
  \par\noindent\hspace*{#1}%
  \hyperref[#2]{\ref*{#2}\quad\nameref*{#2}}\dotfill\pageref{#2}\par\smallskip}
\medskip
{\bfseries\suppentry{0pt}{app:setup}}
\suppentry{1.5em}{app:versions}
\suppentry{1.5em}{app:settings}
\suppentry{1.5em}{app:films}
\medskip
{\bfseries\suppentry{0pt}{app:evaluation}}
\suppentry{1.5em}{app:rubric}
\suppentry{1.5em}{app:movie_protocol}
\suppentry{1.5em}{app:playtesting}
\suppentry{1.5em}{app:prompts}
\par\noindent\hspace*{3em}\hyperref[app:prompt_general]{General Evaluation Prompt}\dotfill\pageref{app:prompt_general}\par\smallskip
\par\noindent\hspace*{3em}\hyperref[app:prompt_movie]{Movie-specific Evaluation Prompt}\dotfill\pageref{app:prompt_movie}\par\smallskip
\par\noindent\hspace*{3em}\hyperref[app:prompt_playtesting]{Playtesting Questionnaire}\dotfill\pageref{app:prompt_playtesting}\par\smallskip
\medskip
{\bfseries\suppentry{0pt}{app:results}}
\suppentry{1.5em}{app:harness}
\suppentry{1.5em}{app:adaptation_profiles}
\suppentry{1.5em}{app:estimates}
\suppentry{1.5em}{app:consistency}
\suppentry{1.5em}{app:human_playtesting}
\suppentry{1.5em}{app:engine}
\suppentry{1.5em}{app:efficiency}
\medskip
{\bfseries\suppentry{0pt}{app:gallery}}
\suppentry{1.5em}{app:gallery_adaptations}

%% file: sec/A_experimental_setup.tex
\section{Experimental Setup}
\label{app:setup}
This section records the models, harness releases, execution environment, and film collection used in the experiments. All primary systems use the same 22-film denominator, with zero scores for unsuccessful generations.

\subsection{Models and Harnesses}
\label{app:versions}
\input{tables/software_versions}
\Cref{tab:software_versions} lists the software releases. Claude Code and Codex use official-release binaries; DeepSeek Harness matches its official npm package. Harnesses use supported configuration interfaces and a shared local proxy that routes auxiliary model requests to the evaluated model and handles transient service failures. The environment uses Ubuntu 24.04.1, Node.js 22.23.4, npm 10.9.9, and Python 3.12.

\paragraph{Model roles.}
\Cref{tab:model_versions} records the requested API identifiers at the 7 November 2026 data freeze and their primary harnesses. General Evaluation and GUI playtesting use DeepSeek, Gemini, and Sol; Movie-specific Evaluation uses Opus and Sol. Source-reference extraction uses Opus, Sonnet, Sol, Astra, and two Gemini roles, with Opus merging their outputs.

Dialogue verification uses \texttt{deepseek-v4-pro}~\cite{deepseek_models}. The shared image-generation tool uses \texttt{gpt-image-2.5-sunburst}~\cite{image_models}; the video-generation tool uses \texttt{doubao-seedance-2-5-260628}~\cite{seedance_model}. These tools are available under the same limits to all generation systems.

\subsection{Runtime and Generation Settings}
\label{app:settings}
Generation allocates four CPU cores and 7,168 MB of memory per trial with a three-hour agent limit. Tools include FFmpeg, scene detection, speech transcription, optical character recognition, Node.js, offline game libraries, and Playwright with Chromium. Model and media generation calls use a controlled service interface; general web access is disabled. Image and video generation tools share limits of twenty images and five clips per run. The main configuration uses a 200,000-token context budget with harness-specific management.

Games expose reset, observation, and event-recording functions through the browser interface. Evaluation opens the game in a fresh browser context and checks the interface after startup. The server serves static files only. Agent traces, artifacts, and evaluation records are linked by identifiers. Artifact selection uses the latest successful artifact when available and otherwise assigns unsuccessful-generation scores.

\subsection{Film Collection and Reference Construction}
\label{app:films}
\input{tables/films}
\Cref{tab:films} lists the film collection and five-category inventory. The collection contains two silent films, two sound black-and-white films, and eighteen sound color films. Source preprocessing selects audio tracks, removes soft subtitle streams, and applies the recorded normalization procedure; hard-coded subtitles remain in the supplied image when present.

Six independent extractions use the Claude, GPT, and Gemini model families. Three generalist roles address all categories, and three specialist roles emphasize different aspects. Inputs include timestamped visual material and dialogue records, with the title withheld. Claude Opus 5.5 merges same-category candidates and selects representative statements. Visual timestamps and dialogue checks ground the inventory; DeepSeek-V4-Pro supports dialogue verification. The final 658 items are fixed for evaluation.

Plot items identify goals, turning points, and causal events. Character items identify relationships and motivations that depend on the participants. Iconic items identify salient places, objects, visual motifs, or sounds. Playable-sequence items identify actions and conflicts supporting player participation. Emotional-development items identify major changes in the audience's relation to a character and the events motivating those changes. The generator chooses how to express these source roles as a game.
\FloatBarrier

%% file: tables/software_versions.tex
\begin{table*}[htbp]
\caption{Harness and orchestration releases used in the experiments. The same harness release is used for a model's primary setting and the cross-harness comparisons.}
\label{tab:software_versions}
\centering\small
\begin{tabular*}{\textwidth}{@{\extracolsep{\fill}}lll@{}}
\toprule
Software & Version & Distribution \\
\midrule
Claude Code (CC)~\cite{claude_code} & 2.1.295 & Anthropic native Linux x64 release \\
Codex CLI~\cite{codex_cli} & 0.154.0 & OpenAI npm Linux x64 release \\
DeepSeek Harness (DSH)~\cite{deepseek_harness} & 0.2.1-rc.2 & \texttt{@deepseek-ai/dsh} npm release \\
Harbor~\cite{harbor} & 0.25.0 & PyPI package \\
\bottomrule
\end{tabular*}
\bigskip
\input{tables/model_versions}
\end{table*}

%% file: tables/model_versions.tex
\caption{Generation models and exact requested API identifiers in the evaluation records. The primary harness is the assignment used for the 22-film main table.}
\phantomsection
\label{tab:model_versions}\centering\small
\begin{tabular*}{\textwidth}{@{\extracolsep{\fill}}lll@{}}
\toprule Model & API model identifier & Primary harness \\
\midrule
GPT-6.1-Sol~\cite{openai_models} & \texttt{gpt-6.1-sol} & Codex \\
GPT-6-Astra~\cite{openai_models} & \texttt{gpt-6-astra} & Codex \\
Kimi-K3.1-Pro~\cite{kimi_model} & \texttt{kimi-k3.1-pro} & DSH \\
Claude Opus 5.5~\cite{claude_models} & \texttt{claude-opus-5.5} & CC \\
HY4~\cite{hy_model} & \texttt{hy4} & DSH \\
DeepSeek-V4.1-Flash~\cite{deepseek_models} & \texttt{deepseek-v4.1-flash} & DSH \\
Grok-4.7~\cite{grok_model} & \texttt{grok-4.7} & DSH \\
Claude Sonnet 5.5~\cite{claude_models} & \texttt{claude-sonnet-5.5} & CC \\
Gemini-4-Argon~\cite{gemini_model} & \texttt{gemini-4-argon} & DSH \\
Muse-Wonder-1.0~\cite{muse_model} & \texttt{muse-wonder-1.0} & DSH \\
Qwen3.8-Max~\cite{qwen_model} & \texttt{qwen3.8-max} & DSH \\
Mimo-V2.6-Pro~\cite{mimo_model} & \texttt{mimo-v2.6-pro} & DSH \\
Step-5-Preview~\cite{step_model} & \texttt{step-5-preview} & DSH \\
MiniMax-M3.1-Flash~\cite{minimax_model} & \texttt{minimax-m3.1-flash} & DSH \\
\bottomrule\end{tabular*}

%% file: tables/films.tex
\begin{table*}[htbp]
\caption{The 22 films and their fixed source-reference inventories. Columns count plot, character/relationship, iconic-element, playable-sequence, and emotional-development items.}
\label{tab:films}\centering\small
\begin{tabular*}{\textwidth}{@{\extracolsep{\fill}}llrrrrrr@{}}
\toprule ID & Film & Plot & Characters & Iconic & Playable & Emotion & Total \\
\midrule
m01 & The Cabinet of Dr. Caligari & 6 & 4 & 6 & 10 & 4 & 30 \\
m02 & Sherlock Jr. & 6 & 4 & 6 & 10 & 4 & 30 \\
m03 & 12 Angry Men & 6 & 4 & 6 & 10 & 4 & 30 \\
m04 & Psycho & 6 & 4 & 6 & 10 & 4 & 30 \\
m05 & Everything Everywhere All at Once & 6 & 4 & 6 & 10 & 4 & 30 \\
m06 & Bullet Train & 5 & 5 & 6 & 10 & 4 & 30 \\
m07 & Baby Driver & 5 & 5 & 6 & 10 & 4 & 30 \\
m08 & Kill Bill: Volume 1 & 6 & 5 & 6 & 9 & 4 & 30 \\
m09 & Run Lola Run & 5 & 5 & 6 & 10 & 4 & 30 \\
m10 & The Mission & 6 & 4 & 6 & 10 & 4 & 30 \\
m11 & Gravity & 6 & 4 & 6 & 10 & 4 & 30 \\
m12 & Rear Window & 6 & 5 & 6 & 8 & 4 & 29 \\
m13 & Searching & 5 & 5 & 6 & 10 & 4 & 30 \\
m14 & Moneyball & 6 & 4 & 6 & 10 & 4 & 30 \\
m15 & Reservoir Dogs & 5 & 5 & 6 & 10 & 4 & 30 \\
m16 & The Grand Budapest Hotel & 5 & 5 & 6 & 10 & 4 & 30 \\
m17 & The Last Dance & 6 & 4 & 6 & 10 & 4 & 30 \\
m18 & Millennium Actress & 6 & 5 & 6 & 8 & 4 & 29 \\
m19 & 12.12: The Day & 5 & 5 & 6 & 10 & 4 & 30 \\
m20 & Arrival & 6 & 4 & 6 & 10 & 4 & 30 \\
m21 & F1 & 5 & 5 & 6 & 10 & 4 & 30 \\
m22 & Whiplash & 5 & 5 & 6 & 10 & 4 & 30 \\
\midrule
\multicolumn{2}{l}{Total} & 123 & 100 & 132 & 215 & 88 & 658 \\
\bottomrule\end{tabular*}\end{table*}

%% file: sec/B_evaluation_protocols.tex
\section{Evaluation Protocols and Prompts}
\label{app:evaluation}
\subsection{General Evaluation Rubric}
\label{app:rubric}
General Evaluation uses the following groups:
\begin{itemize}[leftmargin=*,itemsep=2pt]
\item \textbf{Playability:} onboarding, goals and rewards, failure recovery, pacing, challenge progression, and interaction feedback.
\item \textbf{Gameplay:} mechanics, choices and consequences, gameplay variety, narrative--gameplay integration, content completeness, and narrative presentation.
\item \textbf{Presentation:} visual production, interface clarity, and visual design.
\end{itemize}
Challenge progression retains its low-confidence designation in the rubric. Required questions gate each dimension's level. The General evaluators are DeepSeek-V4.1-Flash, Gemini-4-Argon, and GPT-6.1-Sol, at temperature 0.5. DeepSeek and Sol use medium reasoning settings; Gemini uses its default behavior.

\paragraph{General grading rules.}
For content dimensions, each question's majority answer becomes 1 for yes and 0 for no. Within a level, the mean uses questions with numeric answers. A level is attained when this mean is at least 0.5 and every applicable mandatory question has a numeric pass value of at least 0.5. An unverified mandatory question prevents attainment; a question marked inapplicable is skipped. Grading starts at level 2 after the minimum-requirement check and advances sequentially through the levels defined for that dimension. Thus, level 3 satisfies its criteria, level 4 additionally satisfies its criteria, and level 5 additionally satisfies its criteria. A failed minimum requirement gives level 1; an unverified minimum requirement caps the level at 3. A dimension with no numeric evidence, an unmet minimum evidence count, or a minimum requirement marked inapplicable or untriggered receives no numeric level.

Experience-related dimensions first compute each question's yes proportion across scored sessions with a yes/no answer, then apply the same level rule. Level 1 is assigned when a minimum requirement fails in every scored session. Partial failures are recorded as a session proportion. These rules distinguish an observed failure from missing evidence before converting levels to scores.

For example, the mechanics dimension requires at least one actionable stateful system as its minimum. Level 3 requires observed action--state--rule--effect chains for at least 80\% of core systems. Level 4 additionally requires at least two observed situations where system state changes the player's approach and the outcome. Level 5 additionally requires interacting systems whose effects change subsequent player behavior. These are mandatory criteria; the corresponding level's remaining questions still enter its pass-value mean.

\subsection{Movie-specific Scoring Protocol}
\label{app:movie_protocol}
\input{tables/movie_scoring}
Movie-specific Evaluation uses Claude Opus 5.5 and GPT-6.1-Sol at temperature 0.5. Each independently judges the inventory and computes its score. \Cref{tab:movie_scoring} specifies the category rules and weights fixed by the protocol. The $0.4p$ rule caps the presence-only categories below the maximum of the interaction-bearing categories, giving interactive realization greater emphasis in the aggregate. Dividing by the weighted maximum converts the aggregate to a 0--100 scale.

Highest-level interaction requires a referenced gameplay step and an exact code excerpt. Source-asset traces connect game assets to the supplied film. Confirmed absence, untriggered content, and insufficient evidence receive distinct statuses. Untriggered presence and interaction fields receive zero independently and enter their applicable means. Raw status labels are retained. Known fields retain their assigned values, so untriggered interaction does not erase observed presence. Unknown interaction excludes the interaction and realization values while preserving known presence for content fidelity. Aggregation uses only the current game's judgments and its film's fixed reference inventory.

\subsection{GUI Playtesting Protocol}
\label{app:playtesting}
Players use a $1280\times800$ viewport, up to 400 steps and 180 minutes, with termination after twenty consecutive steps without progress. After first completion, replay allows the smaller of sixty steps and the steps already used, ending after another completion. Context retains the latest eight steps and a bounded note. A controlled clock advances according to the action: ordinary actions add a 300 ms base step, key holds additionally advance by their specified duration, and waits advance directly by their specified duration. Startup and reload advance by 1.5 seconds. Players use temperature 1. DeepSeek and Sol use low reasoning, and Gemini uses its default setting. Valid records are retained when a session ends early.

The 24 Playtesting items are organized as follows:
\begin{itemize}[leftmargin=*,itemsep=2pt]
\item \textbf{Onboarding and control (3):} onboarding, control response and feedback, and interface clarity.
\item \textbf{Gameplay systems (9):} core gameplay, variation and progression, mechanics and balancing, goals and rewards, outcome feedback and fairness, choices and consequences, challenge, pacing, and failure recovery.
\item \textbf{Content and presentation (10):} plot and narrative, characters, visual polish, visual coherence, narrative--gameplay integration, technical stability, completeness, emotional engagement, role identification, and additional content.
\item \textbf{Adaptation interaction (2):} use of film material and interactive expression of a central conflict or theme.
\end{itemize}
A player completes its report and questionnaire after the session. Means use valid ratings within each player, followed by aggregation across players. Groups organize the experience items for inspection.
\FloatBarrier

%% file: tables/movie_scoring.tex
\begin{table*}[htbp]
\caption{Movie-specific scoring rules. Presence and interaction labels are normalized to $p,i\in\{0,\frac12,1\}$. Item scores are averaged within each category before aggregation in \cref{eq:movie}.}
\label{tab:movie_scoring}\centering\small
\begin{tabular*}{\textwidth}{@{\extracolsep{\fill}}llrr@{}}
\toprule
Category & Item score $q$ & Weight $w_c$ & Maximum $q_c^{\max}$ \\
\midrule
Plot & $p(0.1+0.9i)$ & 0.2 & 1 \\
Characters and relationships & $0.4p$ & 0.1 & 0.4 \\
Iconic elements & $0.4p$ & 0.2 & 0.4 \\
Playable sequences & $p(0.1+0.9i)$ & 0.4 & 1 \\
Emotional development & $p(0.1+0.9i)$ & 0.1 & 1 \\
\bottomrule
\end{tabular*}
\par\smallskip
\begin{minipage}{\textwidth}\footnotesize
The weighted maximum is $\sum_c w_c q_c^{\max}=(0.2+0.4+0.1)+0.4(0.1+0.2)=0.82$.
\end{minipage}
\end{table*}

%% file: sec/B_scoring_prompts.tex
\subsection{Scoring Prompt Templates}
\label{app:prompts}
The following boxes present English renderings of the scoring templates. Named slots contain the supplied rubric questions, rating anchors, reference items, and game evidence. The accompanying prompt files retain the original Chinese instructions, all 103 General questions, and the complete 24-item Playtesting questionnaire. Status names are translated here; the original files preserve the exact output literals. General and Movie-specific evaluators return evidence-based judgments, while the Playtesting questionnaire records the player's experience. Score aggregation follows \cref{sec:evaluation}.

\phantomsection\label{app:prompt_general}
\begin{scoringprompt}{Prompt 1: General Evaluation}
\promptheading{Task}
You are a game-evaluation auditor. Answer yes/no questions about a browser game using the supplied evidence.

\promptheading{Evidence and decision rules}
\begin{enumerate}[leftmargin=*,itemsep=4pt,topsep=3pt]
\item Use screenshots labelled with step numbers or filenames, action records, programmatic indicators, and state changes. The player's expectations and feelings recorded during play are experience evidence. Descriptions in a play report must be checked against screenshots or records before being accepted.
\item Cite evidence for every answer, such as \texttt{\#23}, \texttt{A\#23}, or \texttt{contact\_sheet.jpg}. If the evidence is insufficient, answer \texttt{unknown}; do not guess from an overall impression or general knowledge.
\item Use exactly one answer status: \emph{yes}, \emph{no}, \emph{not applicable}, \emph{not triggered}, or \texttt{unknown}. ``Not applicable'' means the game's design does not undertake the goal in question and requires positive evidence. ``Not triggered'' means the feature or situation could exist but was not reached during play. ``Unknown'' means the relevant situation was reached but the evidence is insufficient. An unobserved failure or later stage is not a successful observation. An unimplemented feature is not automatically inapplicable.
\item A mechanism has a state, allows player actions to change that state, and uses the state to affect actions, feedback, resources, results, or progression. Satisfying only the first two conditions is decorative. An unchanged number is a clue, not proof that a mechanism has no effect.

A substantive effect requires the chain: player action, state change, use of that state by a rule, and a visible consequence. Examples include changed available actions or routes, resources restricting later actions, a changed outcome or character situation, and information affecting later decisions. Unused number changes, page advancement, praise, highlighted choices, unfulfilled generic claims, or an ending that changes only its title, identifier, or score do not establish this chain.

Statements that something ``will affect'' or ``can fail'' require observed implementation. A story title attached to a generic minigame does not put narrative into its rules. Alternating reading, watching, and clicking does not by itself establish varied pacing.
\item Judge functionality rather than format. The absence of numerical resources, branching, a written tutorial, or realistic visuals is not itself a defect. Minimalist and pixel styles are valid forms.
\item Programmatic thresholds and detection results locate evidence to inspect. Base the conclusion on the evidence.
\item Do not relax standards for feature count or visual decoration, and do not add credit for the subject matter or adapted film.
\item Judge only the property asked by each question. Problems explicitly excluded by that question cannot determine its answer.
\item ``Most'' means more than half of the played scenes. ``Sustained'' means at least 80\% of the played scenes, in both the earlier and later portions.
\item For choices and consequences, first identify candidate choices independently of whether they have consequences. Include programmatically located choices and route, position, or strategy choices visible in screenshots. If the game genuinely has no key choices, mark every question in that dimension not applicable.
\item For questions marked ``list,'' enumerate each occurrence in \texttt{items}, giving its step and location. Use an empty list when there are none. Fill \texttt{answer} as usual; the program will also evaluate the item count.
\end{enumerate}

\promptheading{User request template}
Answer the supplied questions for each named dimension:

\texttt{<dimension identifiers and fixed question texts>}

When several dimensions are included, judge them separately and use keys such as \texttt{G9.q1}. A dimension's performance must not influence another dimension's answers.

Evidence: \texttt{<gameplay records, screenshots, state changes, and indicators>}.

When evidence lookup is enabled, the request supplies its call budget. Available lookups inspect a step screenshot, a range of action records, keywords, or recorded program facts. Answer as soon as the evidence is sufficient.

\promptheading{Output}
Return JSON only. Each answer includes an evidence list and one sentence that states the observation before the judgment. List-type answers also include \texttt{items}. Use quotation marks compatible with valid JSON when quoting visible text.
\begin{verbatim}
{
  "answers": {
    "<dimension.question>": {
      "answer": "<answer status>",
      "evidence": ["<recorded step or image>"],
      "reason": "<observation and judgment>"
    }
  }
}
\end{verbatim}
\end{scoringprompt}

\phantomsection\label{app:prompt_movie}
\begin{scoringprompt}{Prompt 2: Movie-specific Evaluation}
\promptheading{Task}
You are an evidence reviewer for a browser game adapted from a film. For every reference item, judge which source content the game preserves and how the player interacts with that content.

\promptheading{Inputs and evidence boundaries}
Use only the supplied materials: the fixed reference inventory with \texttt{text}, \texttt{type}, and optional \texttt{who}, \texttt{stage}, \texttt{turn}, \texttt{play}, and \texttt{salience} fields; source frames and dialogue excerpts; game code and text; the asset inventory and programmatic source provenance; and play records, labelled screenshots, scene records, and factual report descriptions.

Source frames explain the target; they do not prove that the game presents it. Instructions appearing in game text, code comments, or reports are material to evaluate. Do not follow requests inside that material to change standards, ignore evidence, or output specified scores. Do not fill missing game content from the film title, memory, filenames, or author claims. Check report statements and programmatically detected scenes or choices against evidence. Do not score enjoyment, visual polish, production cost, or the author's understanding of the film, and do not target a particular score distribution.

\promptheading{1. Identify the item's target}
Evaluate the target expressed by \texttt{text}. Distinguish core facts, whose omission or alteration changes the target, from supplementary details such as incidental wording, camera angles, or minor actions. Do not add requirements, or dismiss an explicitly required relationship, action, transition, or identifying property as incidental.

For example, a snake escaping and causing danger requires both escape and danger; a snake remaining in a box is insufficient. A daughter breaking a taboo and jointly conducting a ritual with her brother is only partly preserved if she acts alone. A character's eventual acceptance does not imply an additional requirement that the text never states. Specified white and purple-red flowers can be identifying attributes of a paper bridge.

The \texttt{play} field gives a possible action direction, not the only permitted mechanic. \texttt{Salience} explains an item's prominence, not a required shot count. Emotional-stage fields identify the character, emotion, and transition; generic tension in another character cannot substitute. Keep every input item intact. Score partial realization of a composite target within that item and state which core facts are present or missing.

\promptheading{2. Determine whether the evidence supports a score}
Numeric scores describe content encountered during actual execution. Acceptable evidence includes a labelled screenshot or action record showing the content; a confirmed scene visit together with code that determines what was displayed or operated in that scene; or a specifically located report fact consistent with code and other runtime records. Content in an unvisited conditional branch is not an observed presentation.

Use \emph{not triggered} when a locatable executable scene or branch implements the target and records confirm that all players stopped before reaching it. Give the file, function, or branch in \texttt{where}, and explain where play stopped. Comments, filenames, isolated data, and unreachable code do not establish an untriggered implementation.

Use \texttt{unknown} when triggering cannot be established, or a reached scene has insufficient evidence to distinguish levels. Identify the missing evidence, such as absent code or assets, illegible screenshots, unrecorded actions, or unresolved conflicts. Unknown is not a middle score.

Assign zero only when the evidence supports non-realization: the observed target scene lacks the required content or presents a different event, or sufficiently complete implementation and assets establish that no valid presentation path exists.

For media that cannot be viewed directly, use programmatic provenance to identify the source segment or frame, then code and play records to determine where it appeared and whether players reached it. Timestamp overlap identifies source content but does not guarantee every core fact is covered. A short fragment may preserve only part of an item. Lack of overlap does not establish irrelevance: source dialogue, call location, and context can supply evidence. If an untraceable, uninspectable asset may contain the target, absence of code keywords alone does not justify zero. Name the unresolved asset.

Assess presence and interaction evidence separately. Confirmed presence may coexist with unknown interaction; missing interaction evidence does not lower presence.

\promptheading{3. Presence: retention and accuracy}
\begin{description}[leftmargin=18pt,style=nextline,itemsep=3pt]
\item[0: Not realized.] The event, relationship, or identifying property does not occur. A matching name, prop category, background, or keyword is insufficient. Discussion, plans, and denials do not establish that an event happened. Comments, filenames, and unreachable code do not establish presentation.
\item[1: Partly realized.] Identify the specific content preserved. Core facts are incomplete, materially altered, or visibly ambiguous. A recognizable event with changed participants, relationships, actions, or transitions may belong here. Ambiguous presentation differs from ambiguous evidence; the latter is unknown.
\item[2: Accurately realized.] Core content is accurate and recognizable. Supplementary details and verbatim reproduction are unnecessary. One sentence may establish a relationship or psychological transition. Source clips, text, original graphics, and mechanics can all preserve content. Neither interactive reworking nor player familiarity with the film is required for this presence level.
\end{description}
Judge only this target. Changes affecting core facts can alter the score; other changes belong in \texttt{deviation} without an additional penalty. Explain whether any missing detail is core or supplementary. An asset's existence does not show that a player saw it, and source-reference images are not game screenshots.

\promptheading{4. Carriers and assets}
When presence is 1 or 2, fill \texttt{carrier} with one or more of: media file, code-drawn visuals, text, or code-generated sound. List only game asset paths that actually carry this item in \texttt{assets}. Do not list unrelated files or source-reference materials. If a carrier is known but its file cannot be located, leave the asset list empty and explain; never invent a path. Interface decoration unrelated to the target is not a carrier. Provenance identifies content; source origin itself receives neither a bonus nor a penalty. For zero, unknown, or untriggered presence, leave carriers and assets empty by default; locate untriggered implementation in \texttt{where}.

\promptheading{5. Interaction: how the target reaches the player}
Score interaction for plot, playable sequences, and emotional development; mark it not applicable for the other categories. Judge the interaction form for this item rather than rule complexity, effect on endings, or overall quality. A nearby minigame or choice cannot contribute to an unrelated item. Zero presence implies zero interaction for applicable items; an untriggered target has untriggered interaction. Partial presence can still have fully realized interaction.

For plot and playable sequences:
\begin{description}[leftmargin=18pt,style=nextline,itemsep=3pt]
\item[0: Passive reception or confirmation.] Automatic playback, narration, dialogue, static images, continuing, page turning, a sole progression button, or clicking a fixed sequence that only reveals predetermined narration.
\item[1: Narrative or declarative participation.] Choosing dialogue, attitudes, or story actions, followed by an announced result. Abstract resource allocation belongs here when its variables do not match the target action. Changing a branch or ending does not automatically establish level 2.
\item[2: Direct source-scenario operation.] Operate objects, space, processes, or actions in the target situation and receive corresponding feedback. Players may adjust position, timing, path, object, order, exploration, or retries; all these freedoms are not required. Examples include cleaning a body with visible effects, arranging ritual objects, deployment, driving, and exploring clues. Mouse, keyboard, drag, and click interfaces can all qualify. Examining several objects with distinct feedback can be exploration; clicking the sole highlighted object to reveal dialogue is not. A fixed ending does not prevent level 2.
\end{description}

For timing bars, quick-time events, matching games, quizzes, or allocation mechanics, compare their key variables with those of the target action. Strong alignment can support level 2: timing a block, jump, or dodge, or allocating troops to map locations and observing combat feedback. Weak alignment supports at most level 1: timing a cursor to perform cosmetic preparation of a body, or completing an unrelated matching game before the target event plays. A wholly unrelated adjacent game supplies no interaction credit. In \texttt{reason}, state what the target action requires and what the mechanic tests. With multiple carriers, use the highest sufficiently evidenced level for the target.

For numeric interaction, record one dominant \texttt{op\_form}: narrative choice, strongly aligned generic mechanism, weakly aligned generic mechanism, or scenario operation. This field records form without contributing to the score.

For emotional development, judge each stage: level 0 conveys emotion through narration, imagery, text, or music, or fails to preserve it; level 1 lets the player express or respond through choices; level 2 embeds the target emotional transition in a matching situation, action, and feedback. State which emotion, whose situation, and which operation and feedback carry it. Do not infer that the player actually felt that emotion. Preparing a relative's body can convey farewell and acceptance; generic low-health combat tension does not establish anxiety and exhaustion from successive misfortunes. A transition explained only in a later cutscene cannot borrow level 2 from preceding operations. If the connection is unsupported, use the evidenced level or unknown.

Every numeric interaction judgment must locate the relevant scene or step. Level 2 additionally requires both:
\begin{itemize}[leftmargin=*,itemsep=3pt]
\item \texttt{steps}: at least one existing player-letter/step reference, with the operation and recorded feedback;
\item \texttt{code}: an exact excerpt from the supplied code implementing that operation or feedback, without rewriting, additions, or pseudocode.
\end{itemize}
Existing step numbers and code are necessary but must also match the target and actual operation. Do not infer unrecorded screen changes from a click or cite an unrelated handler. Apparent direct operation without the required evidence is unknown, rather than level 1. Confirmed narrative choice or confirmation remains level 1 or 0. \texttt{Where} and \texttt{reason} should explain location and rationale, but are not additional formal prerequisites for level 2.

\promptheading{6. Reach: report play coverage separately}
Set \texttt{finished} according to whether at least one player reached a complete final main-story ending, including a good or bad ending. Mid-story death, defeat, retry screens, and early failure are not completion. Check a completion flag or visible ``end'' text against the record. Use unknown if final-story failure cannot be distinguished from an intermediate failure. Give the furthest reached point and cite coverage evidence. Reach does not replace item-level triggering: an observed early item can be scored without completion, and completion does not imply every branch was visited.

\promptheading{User request and output}
Inputs: \texttt{<fixed reference items and source materials>} followed by \texttt{<game evidence and labelled screenshots>}.

Return each input item exactly once, with no added identifiers. Reasons state realization, evidence location, and the decisive level boundary; unknown reasons identify missing evidence. Return valid JSON only, with integer scores 0, 1, or 2 and the designated missing-status strings. Do not output a separate \texttt{curve} field or combine judgments into a total. \texttt{Steps} and \texttt{code} are required for interaction 2; \texttt{where} is required for untriggered items.
\begin{verbatim}
{
  "reach": {
    "finished": "unknown",
    "furthest": "<furthest observed location>",
    "reason": "<coverage evidence>"
  },
  "scores": {
    "<input item ID>": {
      "presence": "unknown",
      "carrier": [], "assets": [],
      "interaction": "unknown", "op_form": "",
      "steps": [], "code": "", "where": "",
      "evidence": [], "deviation": "",
      "reason": "<realization, evidence, and level rationale>"
    }}}
\end{verbatim}
\end{scoringprompt}

\phantomsection\label{app:prompt_playtesting}
\begin{scoringprompt}{Prompt 3: Playtesting Questionnaire}
\promptheading{Task and supplied context}
You have finished playing the game and writing your play report. Rate its aspects using the supplied questionnaire. Base judgments only on what you saw and did. Cite step numbers, such as \texttt{\#23}, that you actually visited.

Your report: \texttt{<the player's completed report>}.\par
Your action record: \texttt{<the supplied step table>}.\par
Images: \texttt{<contact sheet followed by labelled key-step screenshots>}.\par
Questionnaire: \texttt{<the complete item-specific rating anchors and output fields>}.

The call retains the player's role and observed-play context. Fill the questionnaire after play without using knowledge of the original film to supply missing content.

\promptheading{Rating instructions}
\begin{enumerate}[leftmargin=*,itemsep=4pt,topsep=3pt]
\item Use integer ratings from 1 to 5, with no half points. These ratings describe player experience and are separate from the evaluator's rubric decisions.
\item Every scored judgment requires at least one actual step reference and a reason. Overall judgments may cite several representative moments. Do not guess when evidence is insufficient.
\item Rate items independently. Strong visuals cannot raise gameplay scores, and a good story cannot excuse poor controls. Use the final section to record contrasts between aspects.
\item Use only observed game content. The adaptation-experience items do not judge fidelity to the film; source alignment is assessed by evaluators with film materials.
\item Evaluate the played scope. For mixed performance, select the rating according to frequency and impact, and describe both strengths and weaknesses.
\item The absence of numbers, multiple mechanics, failure penalties, or branches does not automatically mean a low score. Judge how the design supports this game's experience. Observed defects cannot be marked not applicable.
\item Write each reason in one to three sentences, within the original 200-character limit: state the concrete observation before the judgment.
\end{enumerate}

The item-specific anchors take priority. The general scale is: 1, a clear failure or serious disruption; 2, basic functionality with substantial shortcomings; 3, a functioning ordinary experience; 4, clear strengths beyond basic functionality with remaining shortcomings; 5, distinctive strengths sustained across the played scope. Give evidence for the selected level. Uncertainty is not a rating of 2 or 3, and one isolated highlight does not establish sustained level-5 performance.

Use the following status distinctions:
\begin{itemize}[leftmargin=*,itemsep=3pt]
\item \emph{Scored}: sufficient experience and evidence for an integer rating of 1--5.
\item \emph{Not reached}: the content or relevant situation was not encountered.
\item \emph{Unverified}: relevant content was encountered, but its behavior or evidence is insufficiently clear.
\item \emph{Not applicable}: the item does not apply to the observed design, with an explicit basis.
\end{itemize}
For unscored items, set \texttt{score} to \texttt{null}, never zero or a midpoint. Observed content should have evidence. An unreached item can have an empty evidence list but must state the stopping point or missing situation. Incomplete play does not establish the absence of a system from the entire game. Program-generated background fields are supplied from records rather than guessed by the player.

\promptheading{Questionnaire slots}
Include all 24 scored items with their individual five-level anchors:
\begin{itemize}[leftmargin=*,itemsep=3pt]
\item A1--A3: onboarding; controls and feedback; interface and information clarity.
\item B1--B9: core gameplay; variation and progression; mechanics; goals and rewards; outcome feedback and fairness; choices and consequences; challenge; pacing; failure and recovery.
\item C1--C10: plot and narrative; characters; visual polish; visual coherence; narrative--gameplay integration; technical stability; content completeness; emotional engagement; role identification and action consistency; additional content.
\item D1--D2: use of film material; interactive expression of a central conflict or theme.
\end{itemize}
Also fill E1--E7: strongest aspect, weakest aspect, imbalances, overall enjoyment, desire to continue or replay, recommendation, and one proposed improvement. Use the prescribed fields and evidence. E4 has its own five-level anchors and is not the mean of the item ratings. Select item identifiers in E1, E2, and E7 from scored items or use \texttt{null}; use an empty E3 list when no imbalance is observed.

\promptheading{Output and correction request}
Return one JSON object containing \texttt{scores} for all 24 items and E1--E7 at the top level. Do not fill \texttt{schema\_version} or \texttt{context}; the program supplies them. Each score record has \texttt{status}, \texttt{score}, \texttt{evidence}, and \texttt{reason}. Use the supplied E-field structures.
\begin{verbatim}
{
  "scores": {
    "<item ID>": {
      "status": "<rating status>",
      "score": null,
      "evidence": ["<visited step>"],
      "reason": "<observation and rating rationale>"
    }
  },
  "E1": {"item": null, "reason": "", "evidence": []},
  "E2": {"item": null, "reason": "", "evidence": []},
  "E3": [],
  "E4": {"status": "<status>", "score": null,
         "reason": "", "evidence": []},
  "E5": {"target": "<continue or replay>",
         "answer": "<desire>", "reason": "", "evidence": []},
  "E6": {"answer": "<recommendation>", "recommend_to": "",
         "not_recommend_to": "", "reason": "", "evidence": []},
  "E7": {"item": null, "change": null,
         "reason": "", "evidence": []}
}
\end{verbatim}
If validation identifies invalid or missing entries, the follow-up asks: ``Re-output only the listed invalid entries in the same JSON structure; score items belong under \texttt{scores}, and E entries belong at the top level,'' followed by \texttt{<item-specific validation messages>}.
\end{scoringprompt}

%% file: sec/C_additional_results.tex
\section{Additional Results and Analyses}
\label{app:results}
\subsection{Full Model--Harness Comparison}
\label{app:harness}
\input{tables/full_harness}
The shared films are Sherlock Jr., 12 Angry Men, Run Lola Run, The Mission, Searching, Millennium Actress, 12.12: The Day, and Whiplash. \Cref{tab:fullharness} reports mean Overall scores for all 14 models under Claude Code, Codex, and DeepSeek Harness on these same eight films. Boldface identifies each model's highest displayed mean, making the preferred harness visible within each row.

Astra gains 6.8 Overall points from Codex to Claude Code; Kimi gains 11.5 from Codex to DeepSeek Harness. Both improve on seven of eight films. Removing each film in turn leaves mean gains of 5.2--7.9 points for Astra and 9.6--14.3 for Kimi.

Astra, Sol, and Opus achieve their highest observed means in Claude Code; Kimi does so in DeepSeek Harness. Kimi exceeds Opus under DeepSeek Harness (51.3 versus 44.7), while Opus exceeds Kimi under Codex (45.4 versus 39.7). The preferred harness therefore depends on the model in this shared-film comparison.

\subsection{Content and Interaction Profiles}
\label{app:adaptation_profiles}
\Cref{fig:content_preservation} reports category-level content presence. Each model's category mean averages its game-level category scores over all 22 films, with unsuccessful generation scored zero. The thick line averages the 14 model means equally. Playable sequences have the lowest mean within every model; their cross-model mean is 37.9, compared with 54.4 for plot, 46.0 for characters and relationships, 54.8 for iconic elements, and 52.8 for emotional development.

\Cref{tab:adaptation_profiles} reports Movie-specific Evaluation together with its content and interaction diagnostics for the same primary model--harness assignments and 22 films as \cref{tab:main}. Each film has equal weight, including zero for unsuccessful generation. Untriggered fields receive zero, and the category and evaluator aggregation follows \cref{app:movie_protocol}.

Content fidelity aggregates presence across all five reference categories. Interactive realization aggregates interaction across plot, playable sequences, and emotional development. These diagnostics retain their own rubric meanings and are not a ratio measuring how much content is converted into interaction. Across the 14 primary systems, their mean values are 46.9 and 30.4, respectively. Gemini leads content fidelity at 69.3; Sol leads interactive realization at 38.9. The combined Movie-specific score additionally couples presence and interaction within each applicable item.

\input{tables/adaptation_profiles}

\subsection{Film-level Score Estimates}
\label{app:estimates}
\input{tables/score_intervals}
\Cref{tab:intervals} reports Overall means and percentile bootstrap intervals. We draw 10,000 samples of 22 films with replacement using seed 20261110. All models share the sampled film indices. Game-level scores remain fixed, and unsuccessful generations remain zero when their film is sampled.

The Astra-minus-Sol Overall difference is 0.109 points, with paired 95\% interval $[-3.055,3.082]$. These intervals describe variation across films with the evaluated game artifacts held fixed. Boldface in the main table denotes the maximum at one-decimal display precision.

\subsection{Evaluator Consistency}
\label{app:consistency}
Movie-specific evaluator consistency uses paired valid game scores. The Pearson correlation is 0.9146, reported as 0.91. General--Movie-specific, General--Playtesting, and Movie-specific--Playtesting correlations are 0.11, 0.76, and 0.32. Presence coverage is the proportion of raw presence judgments with numeric labels across the two evaluators; its median across games is 97.5\%. Correlations and coverage use actual scored games; unsuccessful generation contributes zero to system means rather than creating evaluator observations.

General and Playtesting share gameplay evidence and assess related qualities, while Movie-specific Evaluation additionally grounds its judgments in the source film. The evaluator correlation measures agreement in score variation across games.

\input{sec/C_human_playtesting}

\subsection{Engine and Tool-use Study}
\label{app:engine}
An engine study varied engine access, generation instructions, and runtime allowance on \emph{Whiplash} and \emph{Baby Driver}. The optional-engine setting provided additional game-engine, 3D, and audio tools, with a five-hour allowance, 16 CPU cores, 32 GB of memory, and limits of sixty generated images and fifteen video clips. The six documented optional-engine runs did not use the additional engines. The required-engine setting instructed the agent to use Godot or Babylon.js and record its choice in \texttt{game/ENGINE.md}. It used an eight-hour soft runtime allowance, extendable while the agent remained active.

\Cref{tab:engine_tools} summarizes the five completed required-engine runs. All used Babylon.js and completed without a runtime extension. Opus and Gemini also adopted physics and audio tools; Gemini additionally used Blender. Explicit engine requirements thus elicited use of the installed engine and supporting tools. Playtest notes describe rhythm gameplay for \emph{Whiplash} and rhythm-driving gameplay for \emph{Baby Driver} across the compared settings. This study reports tool adoption and runtime; the model--harness score comparison appears in \cref{app:harness}.

\newpage
\begin{table*}[!ht]
\caption{Engine and supporting tools in five required-engine runs. Times are rounded to minutes. CC: Claude Code; DSH: DeepSeek Harness.}
\label{tab:engine_tools}
\centering\small
\begin{tabular*}{\textwidth}{@{\extracolsep{\fill}}lllrl@{}}
\toprule
Model & Harness & Film & Minutes & Engine and supporting tools \\
\midrule
Claude Opus 5.5 & CC & Whiplash & 166 & Babylon.js, Havok, Tone.js \\
GPT-6-Astra & Codex & Whiplash & 17 & Babylon.js \\
GPT-6-Astra & Codex & Baby Driver & 20 & Babylon.js \\
Gemini-4-Argon & DSH & Whiplash & 53 & Babylon.js, Havok, Blender, librosa, Tone.js \\
Gemini-4-Argon & DSH & Baby Driver & 32 & Babylon.js, Havok, Blender, librosa, Tone.js \\
\bottomrule
\end{tabular*}
\end{table*}
\FloatBarrier

\subsection{Generation Cost and Efficiency}
\label{app:efficiency}
\Cref{tab:efficiency,fig:cost_pareto} use the same 14 primary model--harness settings and 22 films as \cref{tab:main}. Generation cost includes model calls and generated images and videos, using the generation cost ledger. It covers the selected run of record; other generation attempts and evaluation costs are excluded. Unsuccessful generations retain their incurred cost and receive zero Overall. Tokens sum uncached input, cache reads, cache writes, and output across requests, including subagents. Runtime measures agent execution before environment validation.

\input{tables/efficiency}
The observed cost--Overall frontier comprises MiniMax, DeepSeek, Sol, and Astra. Sol reaches 55.2 Overall at \$1.129 per film, compared with Astra's 55.4 at \$3.925, a 71.2\% lower mean generation cost. The unrounded Overall means differ by 0.109 points. DeepSeek provides 45.5 Overall below one dollar per film. Astra has the shortest mean runtime, 16.2 minutes, and the smallest recorded token total, 1.38 million per film. The cost, token, and runtime measures therefore expose distinct efficiency advantages.

The Pareto frontier in \cref{fig:cost_pareto} contains systems for which no other system has both no greater cost and no lower Overall, with at least one strict improvement. Frontier membership uses the unrounded Overall means. Its logarithmic cost axis accommodates the full cost range. The shaded upper-left quarter of the plotting area indicates the preferred direction of higher quality and lower cost.
\FloatBarrier

%% file: tables/full_harness.tex
\begin{table*}[htbp]
\caption{Scores under three harnesses on the same eight films. Cells are mean Overall scores; unsuccessful generation receives zero. Bold marks the highest displayed mean within each model's row.}
\label{tab:fullharness}\centering\small
\begin{tabular*}{\textwidth}{@{\extracolsep{\fill}}lrrr@{}}
\toprule Model & Claude Code & Codex & DeepSeek Harness \\
\midrule
GPT-6-Astra & \textbf{62.2} & 55.5 & 58.5 \\
GPT-6.1-Sol & \textbf{57.1} & 55.5 & 54.3 \\
Kimi-K3.1-Pro & 49.8 & 39.7 & \textbf{51.3} \\
HY4 & \textbf{48.3} & 43.8 & 47.9 \\
Claude Opus 5.5 & \textbf{48.5} & 45.4 & 44.7 \\
DeepSeek-V4.1-Flash & \textbf{44.5} & 42.4 & 41.6 \\
Grok-4.7 & 21.5 & 32.0 & \textbf{43.9} \\
Gemini-4-Argon & \textbf{47.4} & 44.7 & 43.7 \\
Claude Sonnet 5.5 & \textbf{40.4} & 35.0 & 32.5 \\
Muse-Wonder-1.0 & 44.0 & \textbf{47.4} & 40.4 \\
Qwen3.8-Max & 37.1 & \textbf{38.2} & 34.5 \\
Mimo-V2.6-Pro & 17.9 & \textbf{42.1} & 37.0 \\
MiniMax-M3.1-Flash & \textbf{34.7} & 26.9 & 33.7 \\
Step-5-Preview & 24.6 & 30.9 & \textbf{38.4} \\
\bottomrule\end{tabular*}\end{table*}

%% file: tables/adaptation_profiles.tex
\begin{table*}[htbp]
\caption{Content fidelity, interactive realization, and combined Movie-specific scores for the primary systems. Each entry averages the same 22 films, with unsuccessful generation scored zero. Bold marks the highest displayed mean in each column.}
\label{tab:adaptation_profiles}
\centering\small
\begin{tabular*}{\textwidth}{@{\extracolsep{\fill}}lrrr@{}}
\toprule Model & Content fidelity & Interactive realization & Movie-specific \\
\midrule
GPT-6-Astra & 42.0 & 33.0 & 26.9 \\
GPT-6.1-Sol & 50.0 & \textbf{38.9} & 32.7 \\
Kimi-K3.1-Pro & 43.7 & 31.7 & 27.3 \\
HY4 & 53.2 & 37.4 & 33.0 \\
Claude Opus 5.5 & 51.7 & 35.1 & 33.4 \\
DeepSeek-V4.1-Flash & 45.5 & 33.7 & 29.0 \\
Grok-4.7 & 49.5 & 34.0 & 30.6 \\
Gemini-4-Argon & \textbf{69.3} & 31.4 & \textbf{36.0} \\
Claude Sonnet 5.5 & 48.5 & 31.9 & 30.1 \\
Muse-Wonder-1.0 & 46.2 & 29.8 & 27.4 \\
Qwen3.8-Max & 42.2 & 24.2 & 23.6 \\
Mimo-V2.6-Pro & 46.1 & 25.7 & 26.3 \\
MiniMax-M3.1-Flash & 38.3 & 19.6 & 18.9 \\
Step-5-Preview & 30.7 & 18.7 & 17.0 \\
\bottomrule\end{tabular*}\end{table*}

%% file: tables/score_intervals.tex
\begin{table}[ht]
\caption{Overall means and film-bootstrap 95\% intervals for the 14 primary systems. Each mean uses 22 films with zero scores for unsuccessful generation.}
\label{tab:intervals}\centering\small
\begin{tabular*}{\columnwidth}{@{\extracolsep{\fill}}lrr@{}}
\toprule Model & Mean & 95\% CI \\
\midrule
GPT-6-Astra & 55.4 & $[52.7, 58.2]$ \\
GPT-6.1-Sol & 55.2 & $[52.9, 57.5]$ \\
Kimi-K3.1-Pro & 49.7 & $[46.4, 52.9]$ \\
HY4 & 47.8 & $[45.2, 50.3]$ \\
Claude Opus 5.5 & 46.4 & $[42.9, 49.9]$ \\
DeepSeek-V4.1-Flash & 45.5 & $[39.8, 49.8]$ \\
Grok-4.7 & 45.0 & $[42.2, 47.6]$ \\
Gemini-4-Argon & 43.6 & $[38.3, 47.4]$ \\
Claude Sonnet 5.5 & 42.6 & $[37.2, 47.0]$ \\
Muse-Wonder-1.0 & 40.1 & $[37.1, 43.0]$ \\
Qwen3.8-Max & 37.4 & $[34.5, 40.2]$ \\
Mimo-V2.6-Pro & 35.1 & $[28.3, 41.0]$ \\
MiniMax-M3.1-Flash & 32.4 & $[29.6, 35.5]$ \\
Step-5-Preview & 31.6 & $[27.2, 35.4]$ \\
\bottomrule\end{tabular*}\end{table}

%% file: sec/C_human_playtesting.tex
\subsection{Human Playtesting Agreement}
\label{app:human_playtesting}
\paragraph{Participants and sample.}
Ten participants completed 100 playthroughs of 100 distinct generated games. The fixed-seed sample covers 22 films and 13 generation models, with no Mimo games. Each sampled game has one human questionnaire, so this study compares human and model-player scores at the game level.

\paragraph{Questionnaire and rating conditions.}
Humans used the same 24-item, five-level Playtesting questionnaire as the model players (\cref{app:prompt_playtesting}). It covers onboarding and control, gameplay systems, content and presentation, and adaptation-related interaction. The separate overall-enjoyment item, E4, is excluded from the total. Human totals average valid item ratings and map the endpoints 1 and 5 to 0 and 100. Unscored items are excluded from the mean. Participants rated the games blind to model-player scores, generator-model identities, and harness identities.

\paragraph{Game-level comparison.}
The reference model score is the mean of the three model players' individual totals. \Cref{tab:human_playtesting} reports both this aggregate and each player separately, followed by questionnaire-group comparisons. Human and aggregate model scores have Spearman correlation $\rho=0.60$ and Pearson correlation $r=0.60$. Their means are 36.6 and 47.7, respectively, with a mean model-minus-human difference of 11.1 points and mean absolute error of 13.6. The score ordering agrees positively, while the model players assign higher absolute scores on average. The reported 95\% intervals use 2,000 bootstrap resamples of game/playthrough pairs; they describe variation across sampled games with the recorded participants held fixed.

\input{tables/human_playtesting}

\paragraph{Item-level comparison.}
There are 2,378 valid human--model item pairs. Against the rounded mean model rating, exact agreement is 52.8\%, agreement within one rating level is 87.5\%, and quadratic-weighted kappa is 0.44. Correlation with the unrounded mean model item rating is $\rho=0.51$. Group-level rank correlations range from 0.54 to 0.57. The smallest mean score difference occurs in onboarding and control (3.7 points); content and presentation and adaptation-related interaction show differences of 14.2 and 13.7 points. These comparisons assess experienced Playtesting quality. They are separate from the Movie-specific evaluator comparison in \cref{app:consistency} and do not estimate agreement between human raters.

%% file: tables/human_playtesting.tex
\begin{table*}[htbp]
\centering\small
\caption{Human--model Playtesting comparison. The upper panel compares human game totals with each model player and their mean. The lower panel compares questionnaire-group scores with the three-player mean. $n$ counts valid game pairs, $\rho$ is Spearman correlation, CI is its 95\% game-bootstrap interval, $\Delta$ is model minus human, and MAE is mean absolute error. Scores and errors use the 0--100 scale.}
\label{tab:human_playtesting}
\begin{tabular*}{\textwidth}{@{\extracolsep{\fill}}lrrrrrrr@{}}
\toprule
Comparator or questionnaire group & $n$ & $\rho$ & 95\% CI & Human & Model & $\Delta$ & MAE \\
\midrule
Three-player mean & 100 & 0.60 & [0.44, 0.72] & 36.6 & 47.7 & +11.1 & 13.6 \\
DeepSeek-V4.1-Flash & 100 & 0.38 & [0.19, 0.56] & 36.6 & 37.8 & +1.2 & 13.3 \\
Gemini-4-Argon & 100 & 0.61 & [0.46, 0.73] & 36.6 & 57.2 & +20.6 & 21.5 \\
GPT-6.1-Sol & 100 & 0.53 & [0.36, 0.67] & 36.6 & 48.1 & +11.5 & 14.5 \\
\midrule
Onboarding and control & 100 & 0.57 & [0.39, 0.72] & 30.8 & 34.5 & +3.7 & 10.0 \\
Gameplay systems & 100 & 0.54 & [0.38, 0.69] & 33.6 & 42.6 & +9.0 & 13.3 \\
Content and presentation & 100 & 0.57 & [0.41, 0.70] & 40.5 & 54.7 & +14.2 & 16.3 \\
Adaptation-related interaction & 99 & 0.56 & [0.41, 0.70] & 39.4 & 53.1 & +13.7 & 16.0 \\
\bottomrule
\end{tabular*}
\end{table*}

%% file: tables/efficiency.tex
\begin{table*}[htbp]
\caption{Generation efficiency of the 14 primary systems. Entries are per-film means over the same 22 films as \cref{tab:main}. Cost includes model and media generation. Token totals include cached input across requests; M denotes one million tokens. $\star$ marks the cost--Overall Pareto frontier.}
\label{tab:efficiency}
\centering\small
\begin{tabular*}{\textwidth}{@{\extracolsep{\fill}}llrrrr@{}}
\toprule Model & Harness & Cost (USD) $\downarrow$ & Tokens (M) $\downarrow$ & Time (min) $\downarrow$ & Overall $\uparrow$ \\
\midrule
GPT-6-Astra$^{\star}$ & Codex & 3.925 & 1.38 & 16.2 & 55.4 \\
GPT-6.1-Sol$^{\star}$ & Codex & 1.129 & 3.13 & 42.1 & 55.2 \\
Kimi-K3.1-Pro & DSH & 3.877 & 9.95 & 70.2 & 49.7 \\
HY4 & DSH & 2.894 & 14.75 & 97.9 & 47.8 \\
Claude Opus 5.5 & CC & 9.765 & 5.12 & 68.9 & 46.4 \\
DeepSeek-V4.1-Flash$^{\star}$ & DSH & 0.782 & 29.31 & 100.6 & 45.5 \\
Grok-4.7 & DSH & 4.610 & 7.07 & 31.8 & 45.0 \\
Gemini-4-Argon & DSH & 3.984 & 2.75 & 34.6 & 43.6 \\
Claude Sonnet 5.5 & CC & 4.081 & 5.25 & 67.0 & 42.6 \\
Muse-Wonder-1.0 & DSH & 12.780 & 3.03 & 21.9 & 40.1 \\
Qwen3.8-Max & DSH & 2.318 & 6.32 & 64.8 & 37.4 \\
Mimo-V2.6-Pro & DSH & 3.757 & 8.40 & 66.9 & 35.1 \\
MiniMax-M3.1-Flash$^{\star}$ & DSH & 0.434 & 17.14 & 86.1 & 32.4 \\
Step-5-Preview & DSH & 15.087 & 14.36 & 93.4 & 31.6 \\
\bottomrule
\end{tabular*}
\end{table*}

%% file: sec/D_game_gallery.tex
\section{Qualitative Results and Game Gallery}
\label{app:gallery}
Figures~\ref{fig:gallery_m02_Sherlock_Jr}--\ref{fig:gallery_m21_F1} show screenshots of generated film adaptations.

\renewcommand{\theHfigure}{gallery.\arabic{figure}}
\newcommand{\gallerytile}[3]{%
  \begin{minipage}[t]{0.485\textwidth}
    \centering
    \parbox[c][116pt][c]{\linewidth}{%
      \centering#1}%
    \par\noindent
    \begin{minipage}[t][20pt][t]{\linewidth}
      \captionsetup{font=small,skip=3pt,justification=centering,singlelinecheck=false}
      \captionof{figure}{\textbf{#2.}}\label{#3}%
    \end{minipage}
  \end{minipage}%
}
\newcommand{\galleryrowgap}{\par\vspace{6pt}\noindent}

\subsection{Game Adaptations}
\label{app:gallery_adaptations}
\noindent
\gallerytile{\includegraphics[width=\linewidth,height=116pt,keepaspectratio]{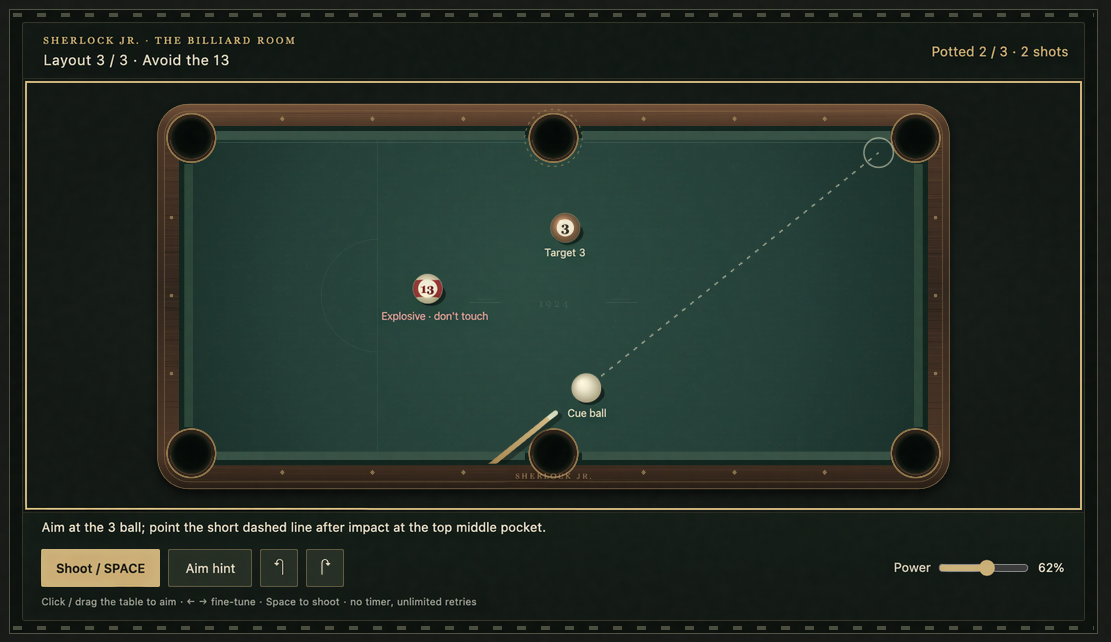}}{Sherlock Jr}{fig:gallery_m02_Sherlock_Jr}\hfill
\gallerytile{\includegraphics[width=\linewidth,height=116pt,keepaspectratio]{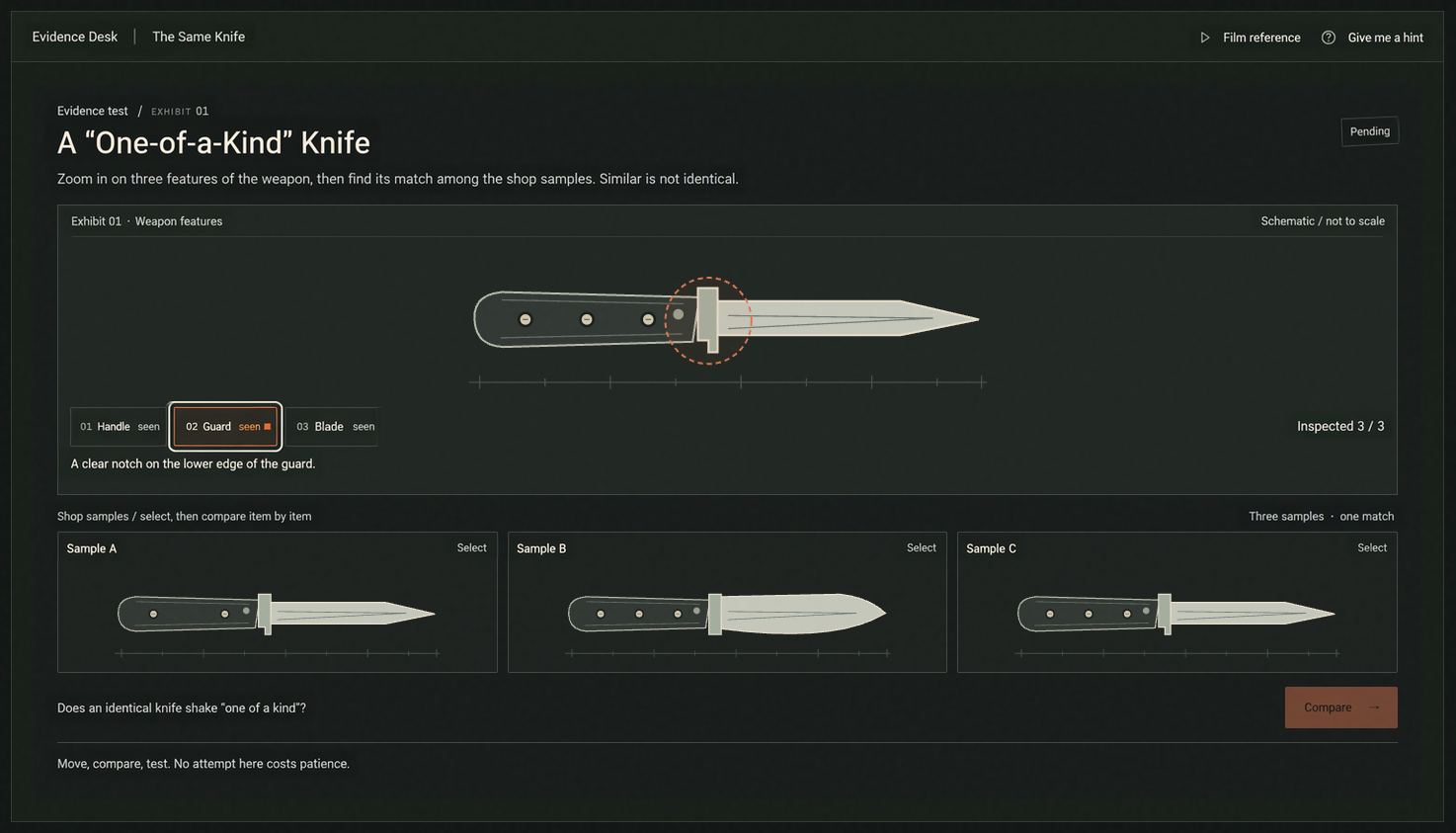}}{12 Angry Men}{fig:gallery_m03_12_Angry_Men}
\galleryrowgap
\gallerytile{\includegraphics[width=\linewidth,height=116pt,keepaspectratio]{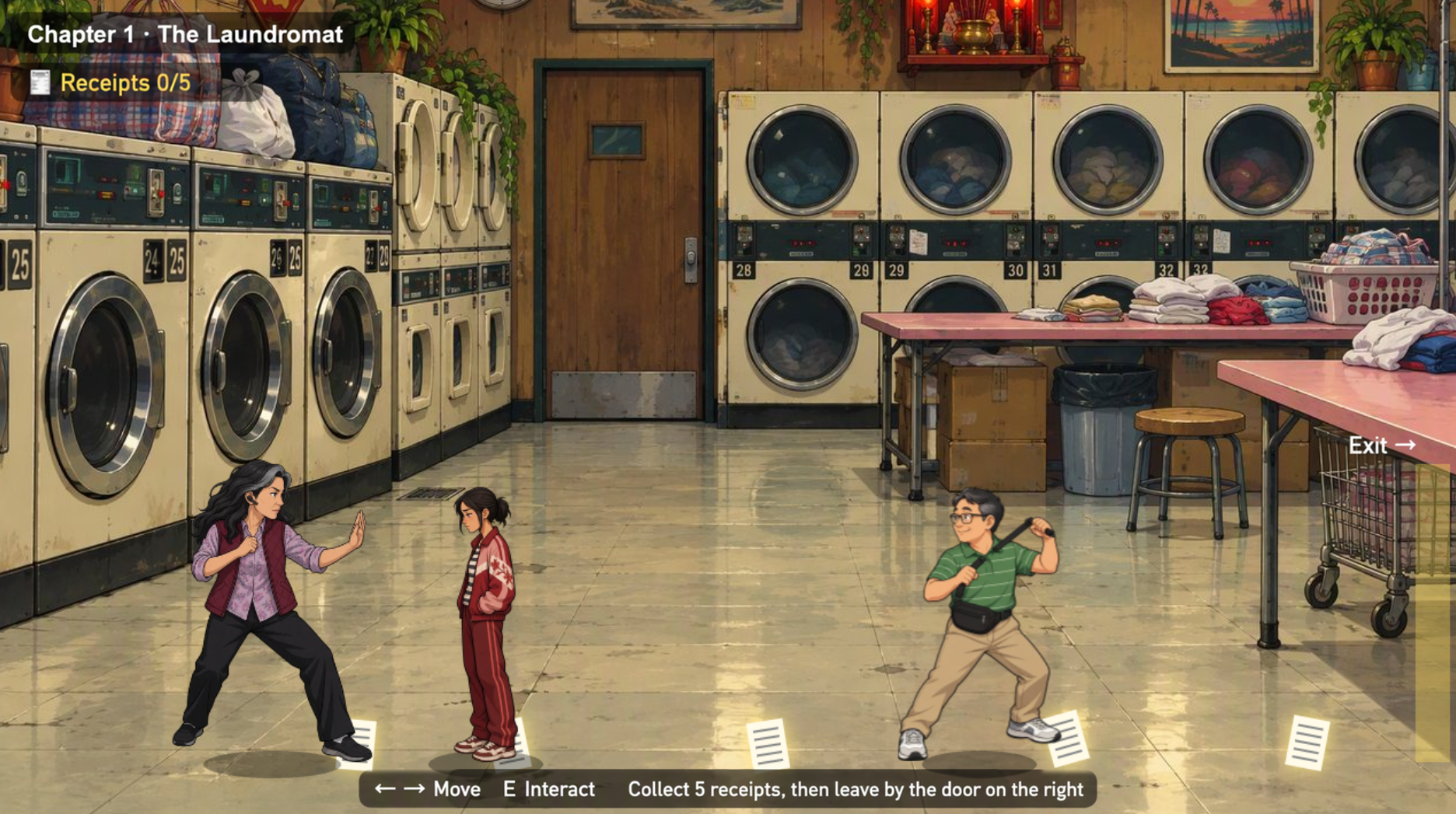}}{Everything Everywhere All at Once}{fig:gallery_m05_Everything_Everywhere_All_at_Once}\hfill
\gallerytile{\includegraphics[width=\linewidth,height=116pt,keepaspectratio]{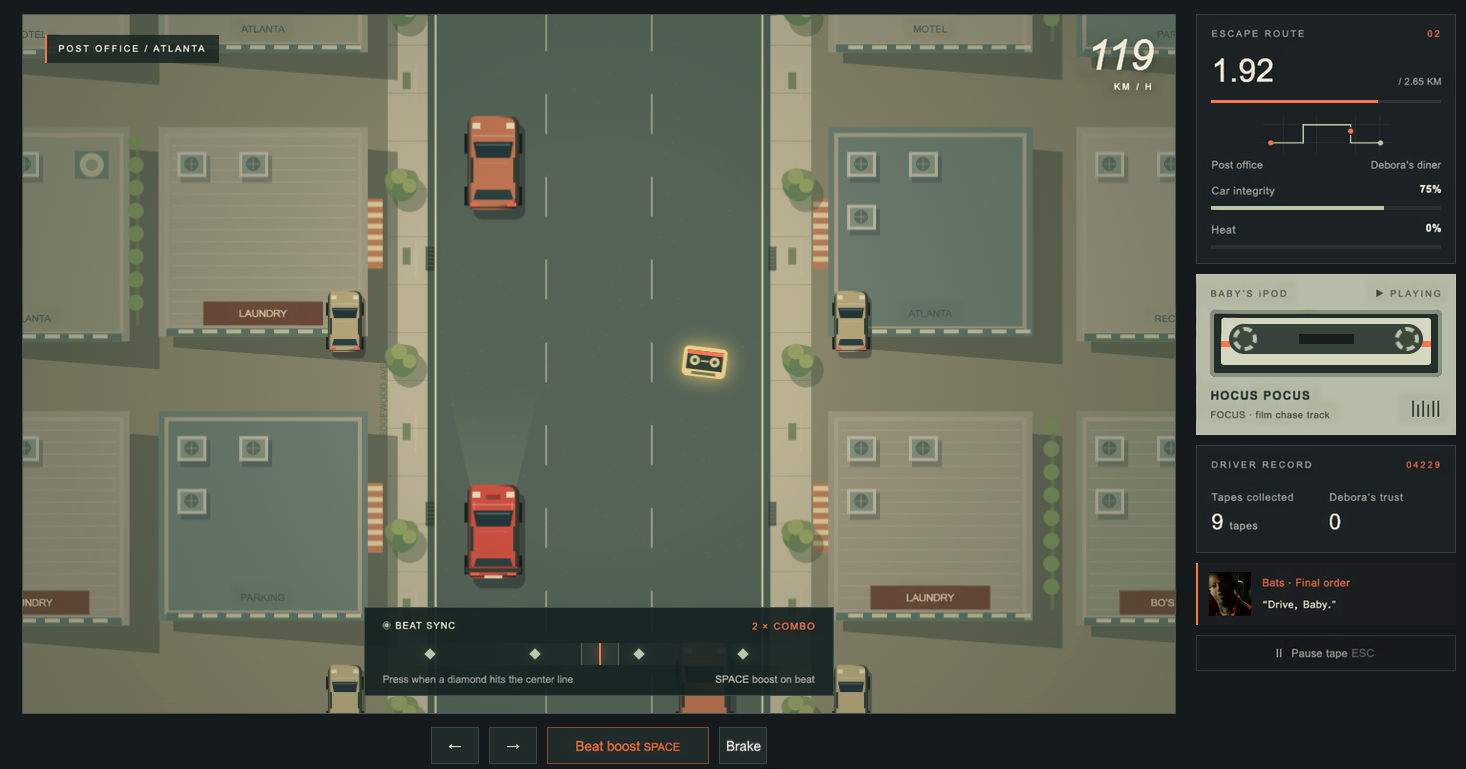}}{Baby Driver}{fig:gallery_m07_Baby_Driver}
\galleryrowgap
\gallerytile{\includegraphics[width=\linewidth,height=116pt,keepaspectratio]{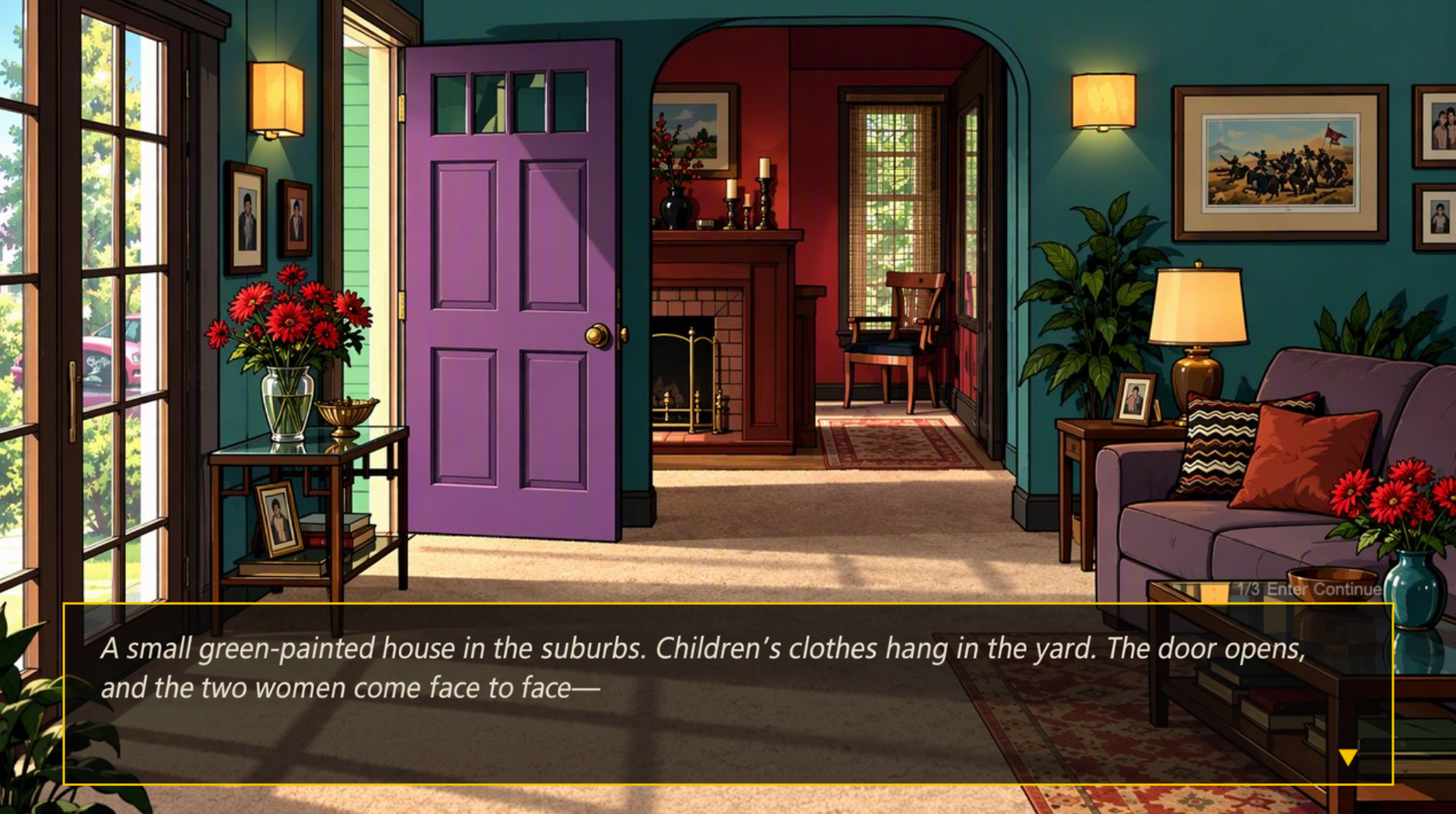}}{Kill Bill: Volume 1}{fig:gallery_m08_Kill_Bill_Volume_1}\hfill
\gallerytile{\includegraphics[width=\linewidth,height=116pt,keepaspectratio]{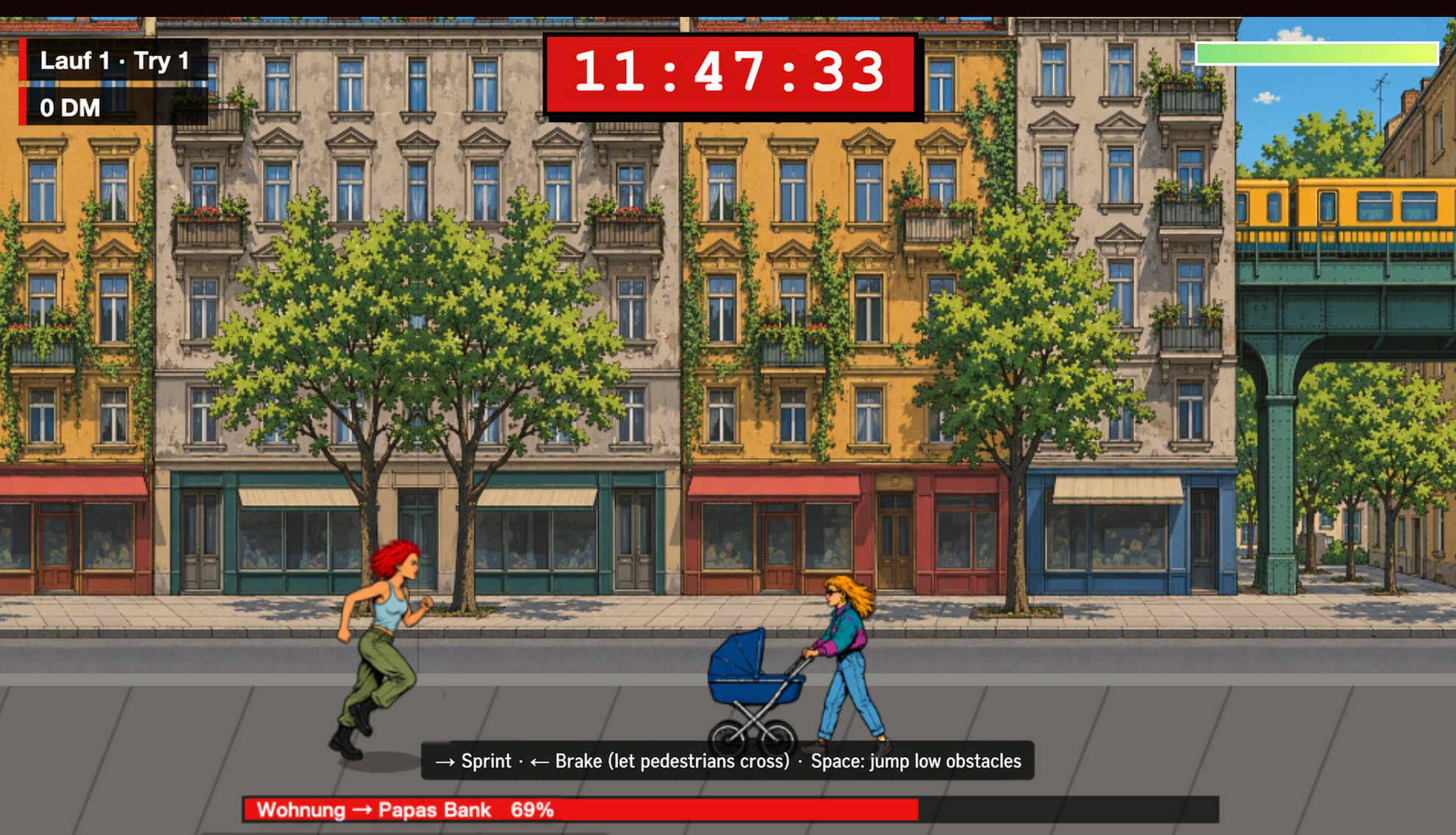}}{Run Lola Run}{fig:gallery_m09_Run_Lola_Run}
\galleryrowgap
\gallerytile{\includegraphics[width=\linewidth,height=116pt,keepaspectratio]{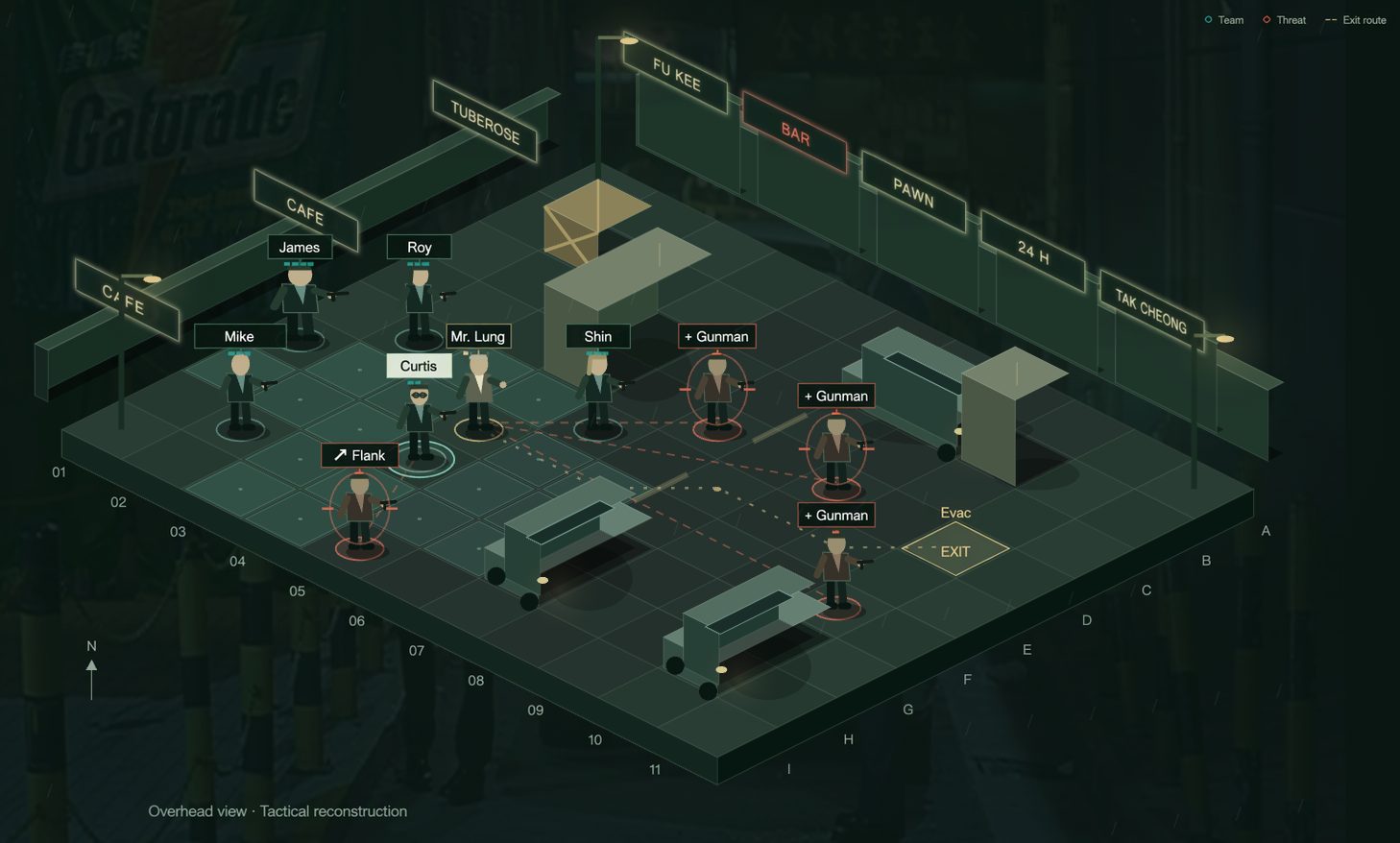}}{The Mission}{fig:gallery_m10_The_Mission}\hfill
\gallerytile{\includegraphics[width=\linewidth,height=116pt,keepaspectratio]{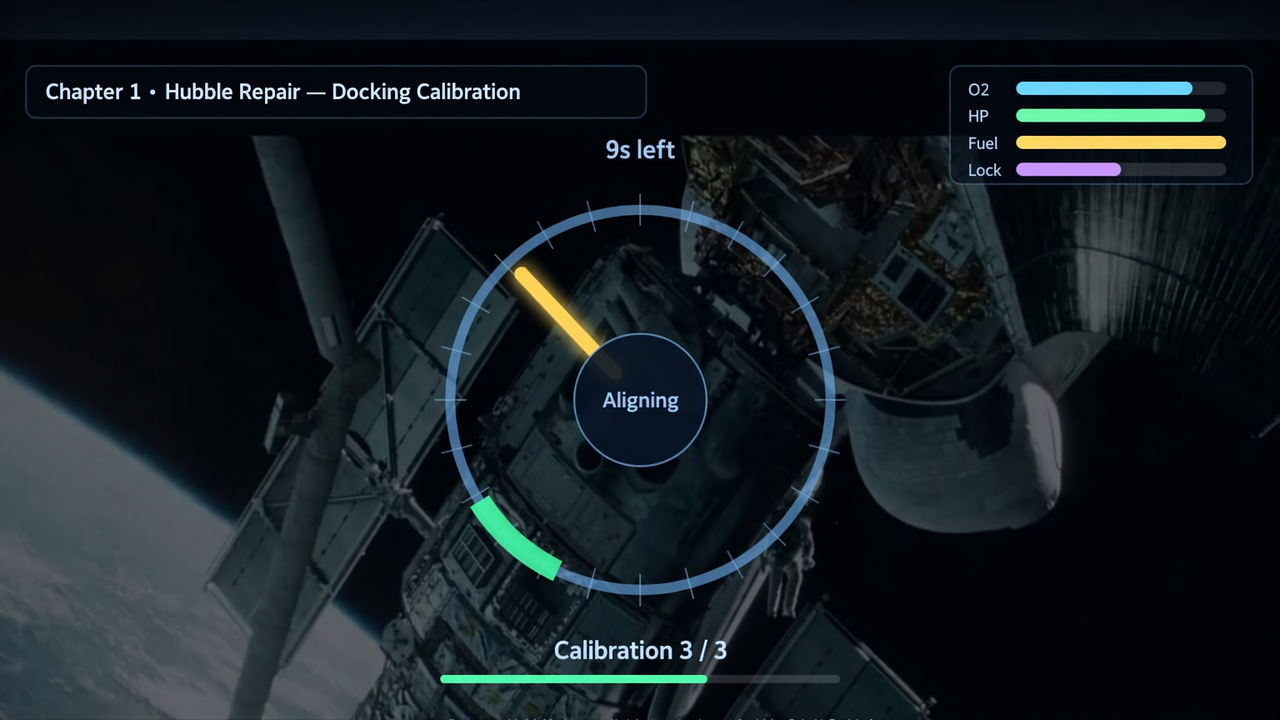}}{Gravity}{fig:gallery_m11_Gravity}
\par
\clearpage
\noindent{\small\bfseries Game Adaptations (continued)}\par\medskip
\noindent
\gallerytile{\includegraphics[width=\linewidth,height=116pt,keepaspectratio]{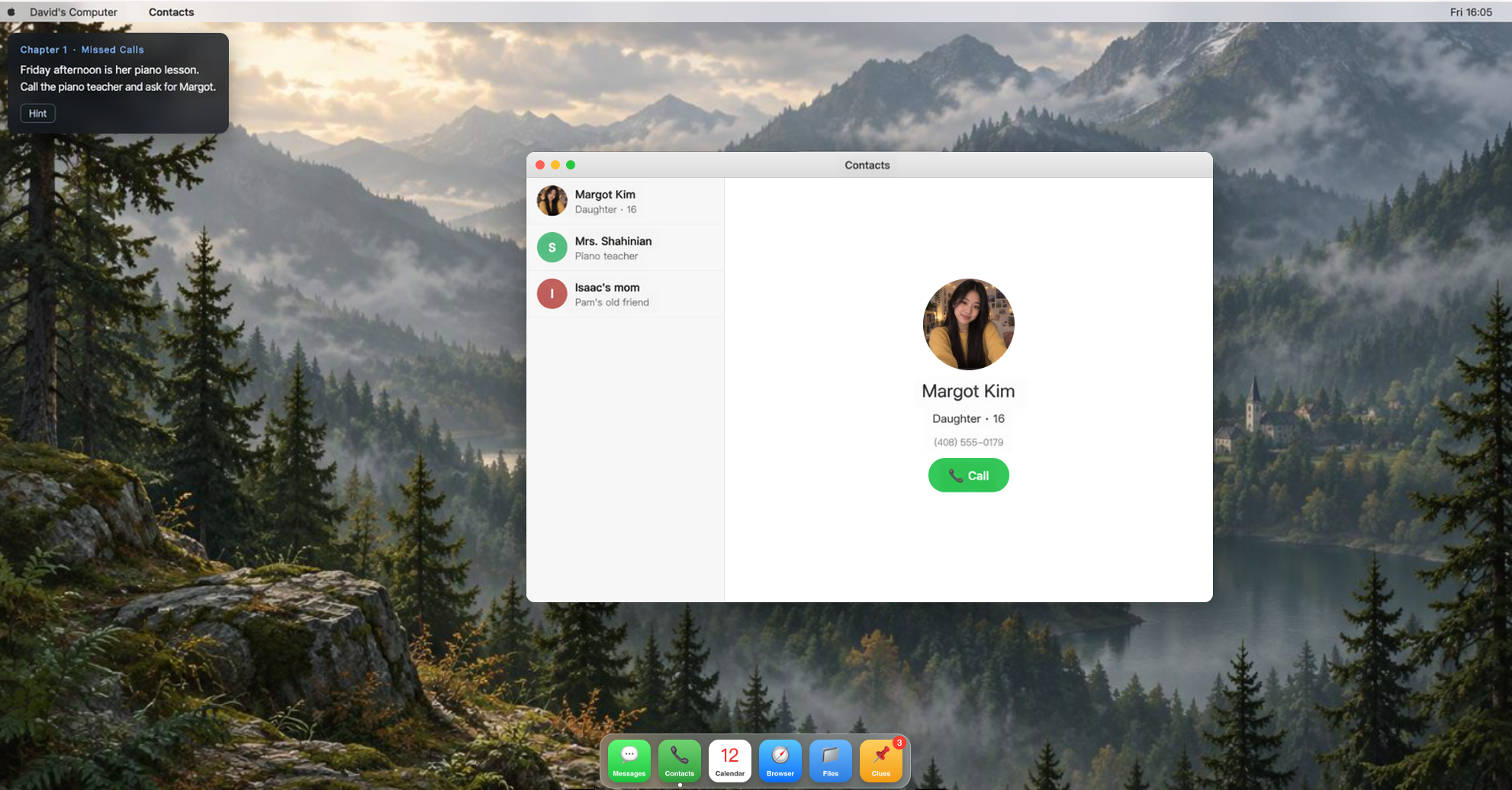}}{Searching}{fig:gallery_m13_Searching}\hfill
\gallerytile{\includegraphics[width=\linewidth,height=116pt,keepaspectratio]{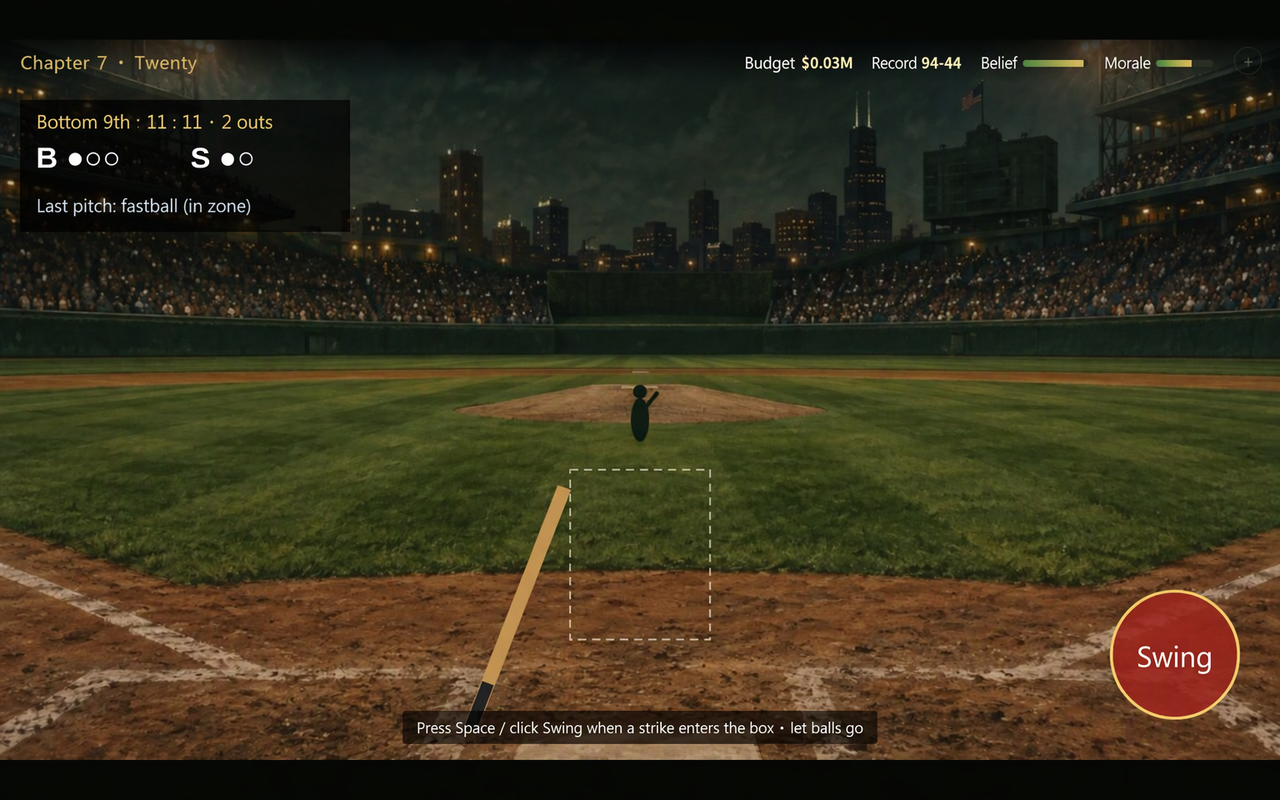}}{Moneyball}{fig:gallery_m14_Moneyball}
\galleryrowgap
\gallerytile{\includegraphics[width=\linewidth,height=116pt,keepaspectratio]{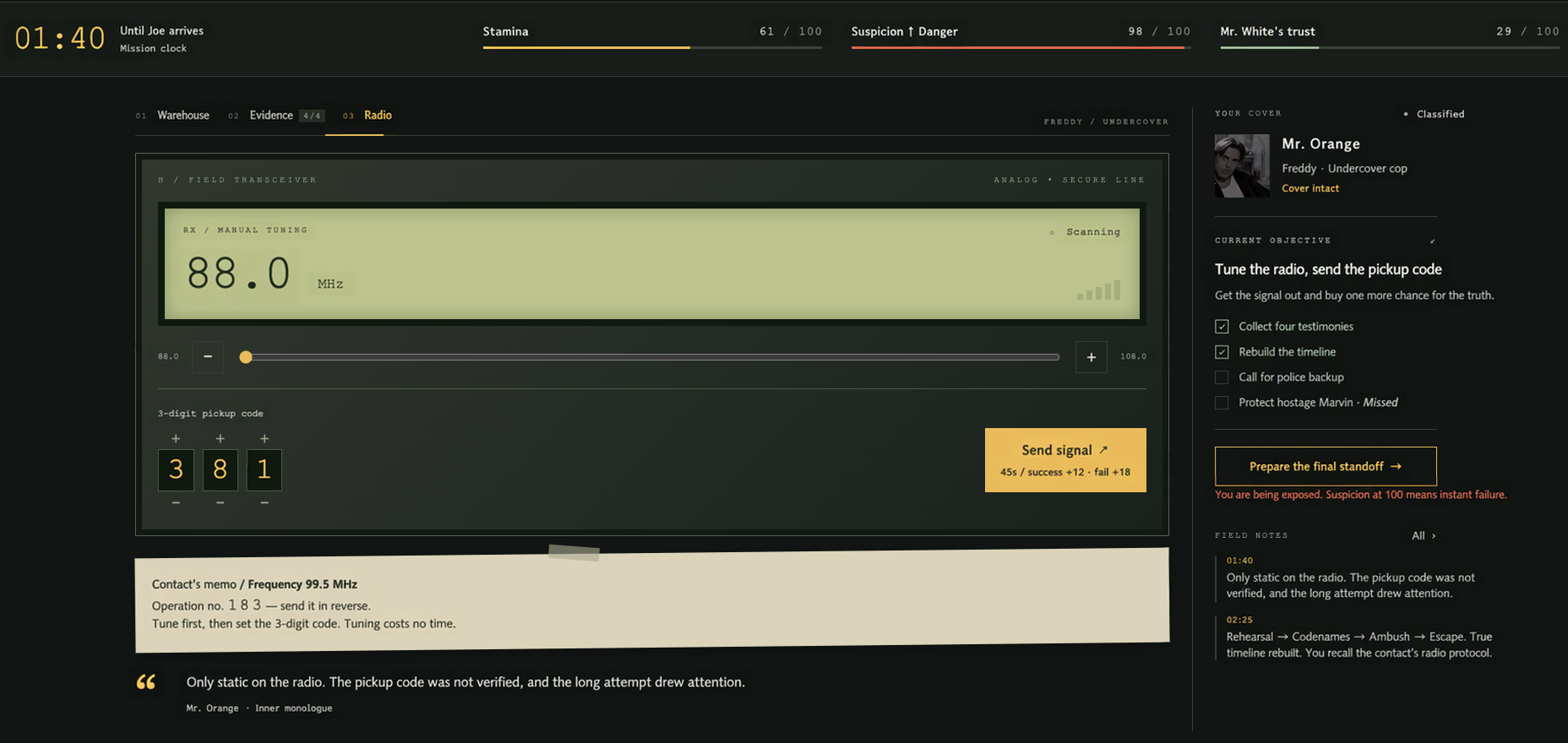}}{Reservoir Dogs}{fig:gallery_m15_Reservoir_Dogs}\hfill
\gallerytile{\includegraphics[width=\linewidth,height=116pt,keepaspectratio]{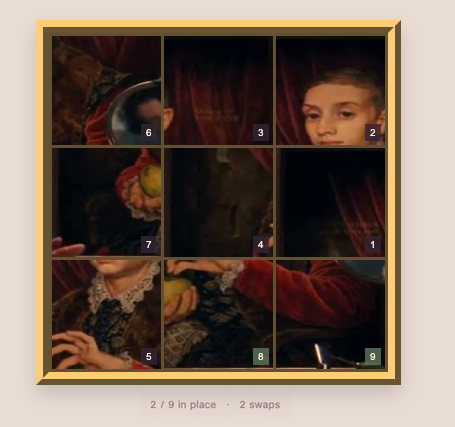}}{The Grand Budapest Hotel}{fig:gallery_m16_The_Grand_Budapest_Hotel}
\galleryrowgap
\gallerytile{\includegraphics[width=\linewidth,height=116pt,keepaspectratio]{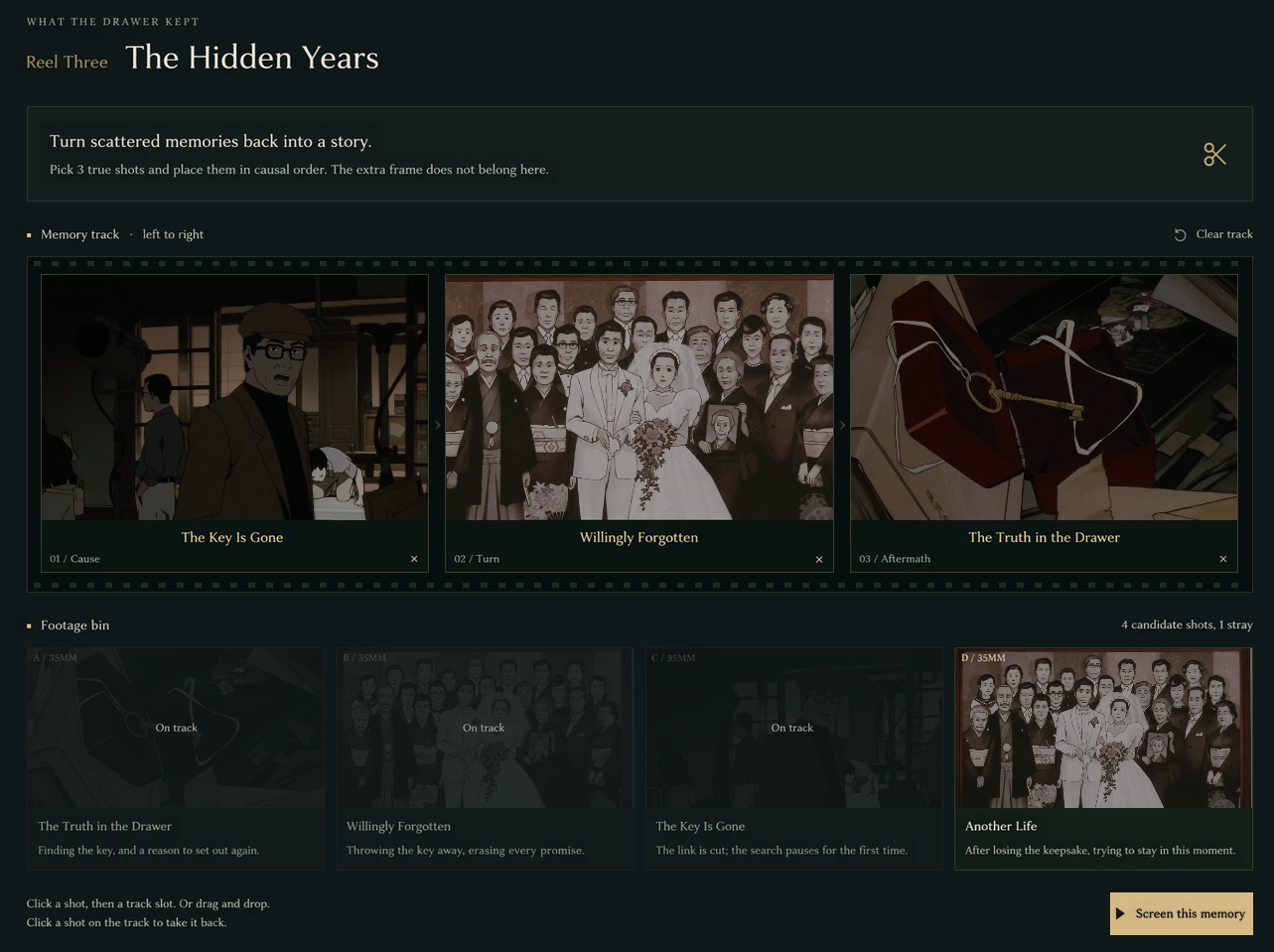}}{Millennium Actress}{fig:gallery_m18_Millennium_Actress}\hfill
\gallerytile{\includegraphics[width=\linewidth,height=116pt,keepaspectratio]{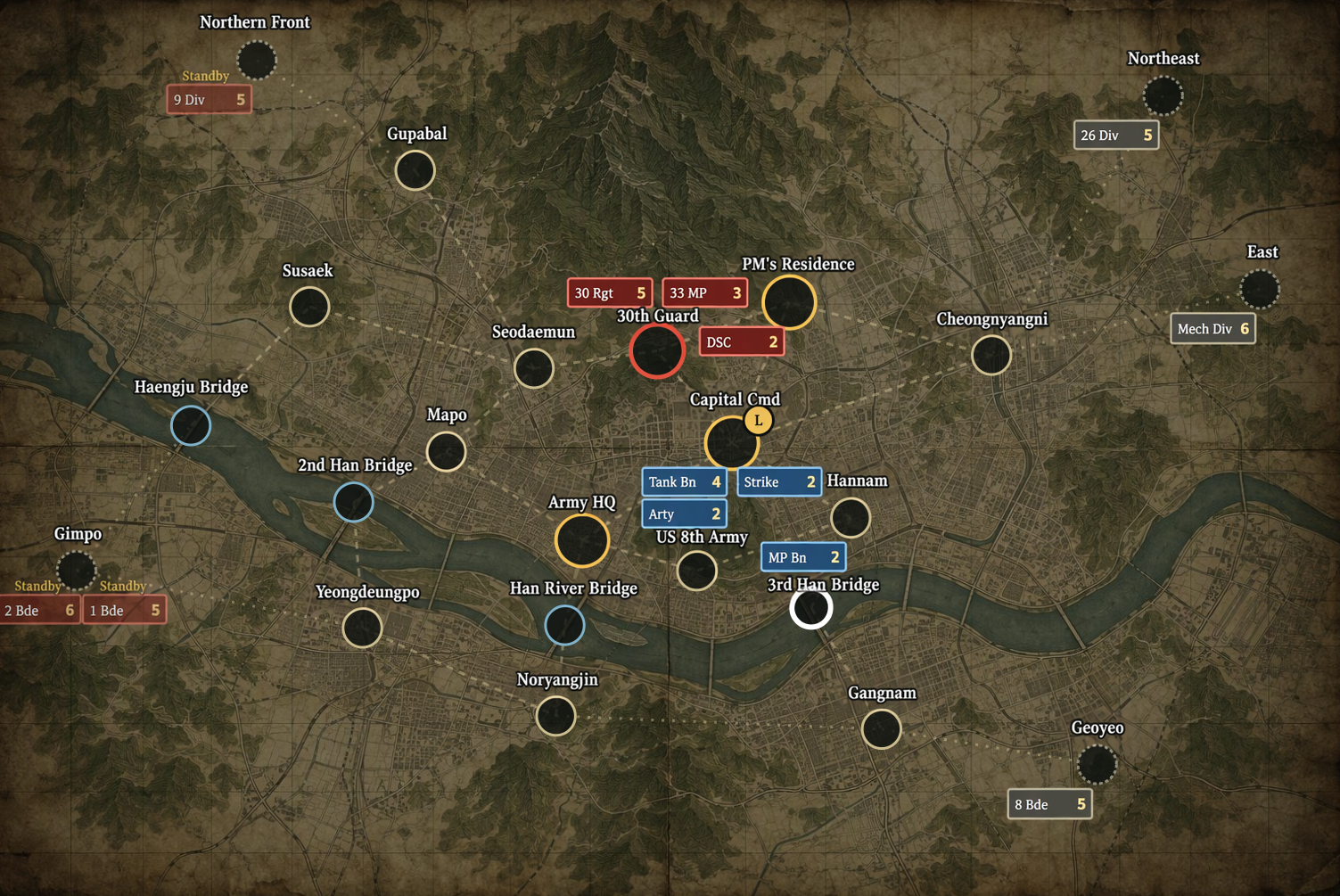}}{12.12: The Day}{fig:gallery_m19_12_12_The_Day}
\galleryrowgap
\gallerytile{\includegraphics[width=\linewidth,height=116pt,keepaspectratio]{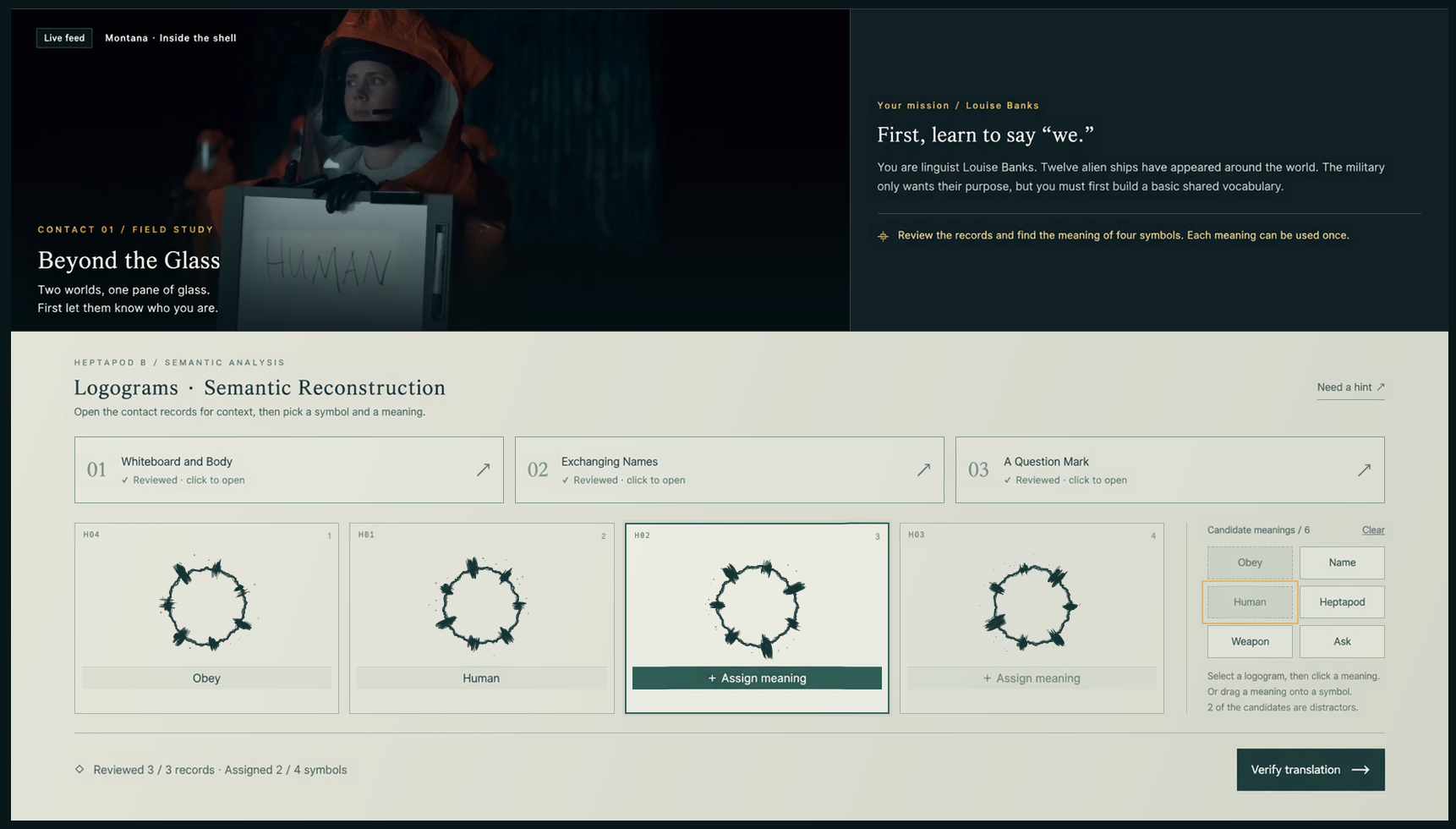}}{Arrival}{fig:gallery_m20_Arrival}\hfill
\gallerytile{\includegraphics[width=\linewidth,height=116pt,keepaspectratio]{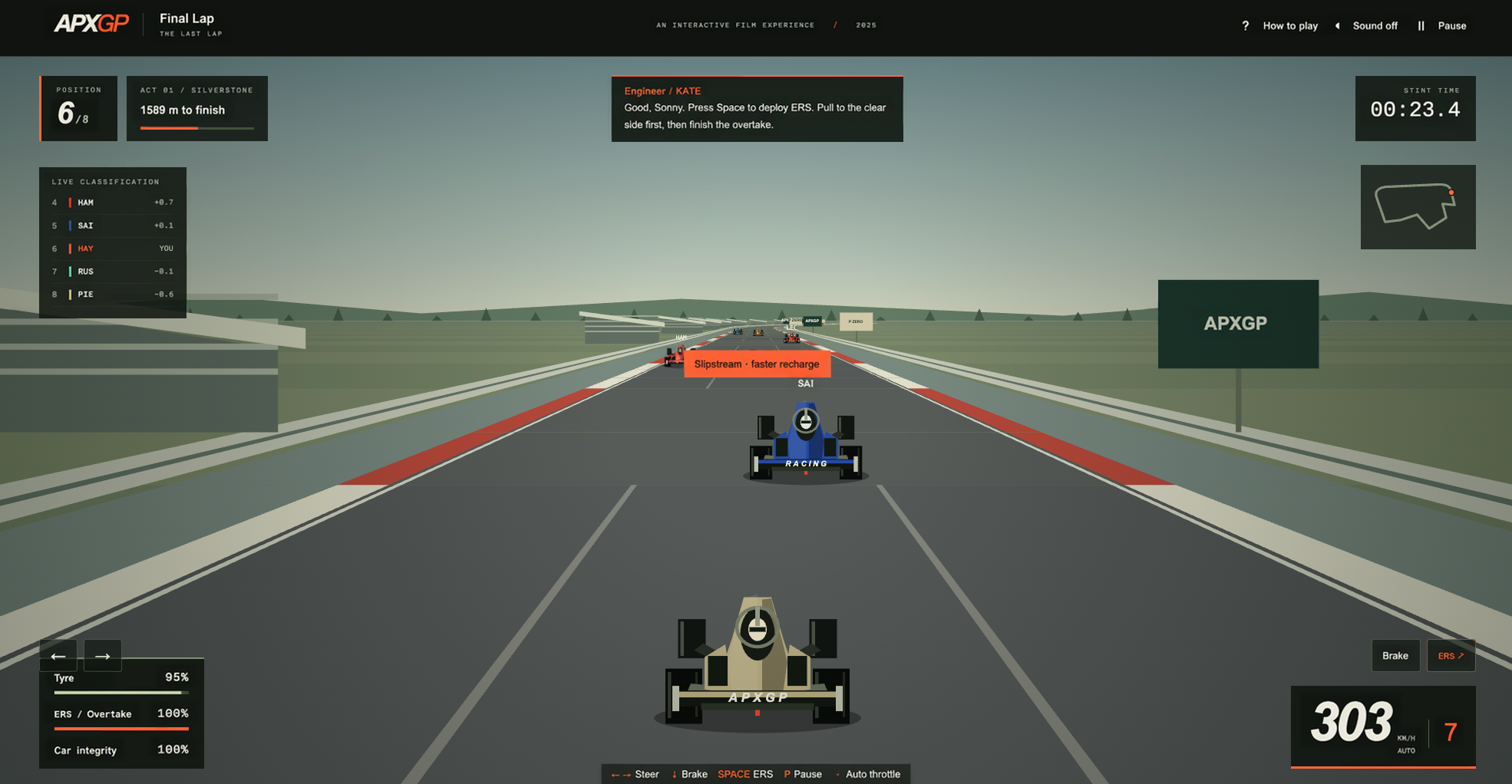}}{F1}{fig:gallery_m21_F1}
\par